\documentclass[journal]{IEEEtran}
\usepackage{hyperref}
\usepackage{amssymb}
\usepackage{subfigure}

\usepackage[table]{xcolor}
\usepackage{longtable}
\usepackage{booktabs}

\usepackage{lineno}
\usepackage{booktabs}
\usepackage{threeparttable}
\usepackage{setspace}
\usepackage{subfigure}
\usepackage{makecell}
\usepackage{rotating}
\usepackage{multirow}
\usepackage{graphicx}
\usepackage{verbatim}
\usepackage{amsmath}
\usepackage{longtable}
\usepackage[T1]{fontenc}
\usepackage[latin1]{inputenc}

\usepackage[explicit]{titlesec}
\titlespacing*{\section}{0pt}{1.2ex plus .0ex minus .0ex}{.3ex plus .0ex}
\titlespacing*{\subsection}{0pt}{1.2ex plus .0ex minus .0ex}{.3ex plus .0ex}
\titlespacing*{\subsubsection}{0pt}{1.2ex plus .0ex minus .0ex}{.3ex plus .0ex}
\usepackage{cases}
\makeatletter
\newif\if@restonecol
\makeatother

\usepackage[ruled,linesnumbered]{algorithm2e}
\usepackage{rotating}
\usepackage{url}
\ifCLASSINFOpdf
\else
\fi

\begin{document}
%
%\linenumbers

\title{Learning-guided iterated local search for the minmax multiple traveling salesman problem}
\author{Pengfei~He,
        Jin-Kao~Hao*,
        and Jinhui Xia
%\thanks{Work partially supported by the National Natural Science Foundation of China (Grant 61903144, the Shanghai Sailing Program (Grant 19YF1412400), the Key Project of Science and Technology Innovation 2030 Supported by the Ministry of Science and Technology of China (Grant 2018AAA0101302), the Fundamental Research Funds for the Central Universities of China (Grant 222201817006), and the Shenzhen Science and Technology Innovation Commission (Grant JCYJ20180508162601910), the funding from Shenzhen Institute of Artificial Intelligence and Robotics for Society (Grant 2019-INT003). (Corresponding authors: Jin-Kao Hao and Zhang-Hua Fu)}
\thanks{P.~He is with the School of Automation and Control of Complex Systems of Engineering, Ministry of Education, Southeast University, Nanjing 210096, China (E-mail: pengfeihe606@gmail.com).}
\thanks{J.-K.~Hao (corresponding authors) is with the Department of Computer Science, LERIA, Universit$\acute{e}$ d'Angers, 2 Boulevard Lavoisier, 49045 Angers, France (E-mail: jin-kao.hao@univ-angers.fr).}
\thanks{J.~Xia is with the School of Automation and Control of Complex Systems of Engineering, Ministry of Education, Southeast University, Nanjing 210096, China (E-mail: jinhuixia@seu.edu.cn).}

%\thanks{Z.-H.~Fu is with the Shenzhen Institute of Artificial Intelligence and Robotics for Society, and The Chinese University of Hong Kong, Shenzhen, 518172 Shenzhen, China (e-mail: fuzhanghua@cuhk.edu.cn).}
%\thanks{X.~Lai is with the Institute of Advanced Technology, Nanjing University of Posts and Telecommunications, 210023 Nanjing, China (e-mail: laixiangjing@gmail.com).}\\
%\thanks{* Corresponding authors.}
\vspace{-0.80cm}
}

%\markboth{IEEE TRANSACTIONS ON }%

\maketitle

\begin{abstract}
The minmax multiple traveling salesman problem involves minimizing the longest tour among a set of tours. The problem is of great practical interest because it can be used to formulate several real-life applications. To solve this computationally challenging problem, we propose a leaning-driven iterated local search approach that combines an aggressive local search procedure with a probabilistic acceptance criterion to find high-quality local optimal solutions and a multi-armed bandit algorithm to select various removal and insertion operators to escape local optimal traps. Extensive experiments on 77 commonly used benchmark instances show that our algorithm achieves excellent results in terms of solution quality and running time. In particular, it achieves 32 new best-known results and matches the best-known results for 35 other instances. Additional experiments shed light on the understanding of the composing elements of the algorithm.

\end{abstract}


% \fi
\IEEEpeerreviewmaketitle
\section{Introduction}\label{intro}

The minmax multiple traveling salesmen problem (minmax mTSP) is formulated within a complete, undirected graph $\mathcal{G} = (\mathcal{V}, \mathcal{E})$, where $\mathcal{V} = \{v_0\} \cup \mathcal{N}$ is the set of vertices, with the vertex $v_0$ serving as the depot and $\mathcal{N} = \{v_1, v_2, \cdots, v_n\}$ being the set of cities, while $\mathcal{E}$ represents the set of edges. Each edge $e_{ij} \in \mathcal{E}$ is associated with a cost $c_{ij}$ and the edge costs satisfy the triangle inequality such that $c_{ij} + c_{jk} > c_{ik}$ for all distinct $v_i, v_j, v_k \in \mathcal{V}$. Given a set of salesmen $\mathcal{M}=\{1,2,\cdots,m\}$, the minmax mTSP aims to find $m$ mutually exclusive Hamiltonian tours (also called routes) such that each tour that starts and ends at the depot $v_0$ must visit at least one city, and each city must be visited exactly once. The objective of the minmax mTSP is to minimize the longest tour among the $m$ tours. A flow-based formulation of the minmax mTSP is shown in the appendix. 

The minmax mTSP, proposed by Fran{\c{c}}a et al. \cite{francca1995m}, has a number of practical applications, where it is necessary to make a fair and equitable distribution of workloads. Typical applications include two-dimensional laser cutting paths \cite{dewil2016review}, multiple robot spot welding paths \cite{zhou2022multi} and patrol and delivery services \cite{zhang2018fast}. Despite its relevance, the minmax mTSP has received much less attention compared to its counterpart, the minsum mTSP, which focuses on minimizing the cumulative traveling cost over $m$ tours. It is worth noting that recently, He and Hao \cite{he2022hybrid} showed that, by transforming the minsum mTSP into the classical traveling salesman problem (TSP) \cite{rao1980note}, the problem can be effectively solved by state-of-the-art TSP algorithms. However, the situation is quite different when it comes to the minmax mTSP. In fact, due to the inherent strong $\mathcal{NP}$-hardness, effectively solving the minmax mTSP is still a computational challenge, despite past efforts on various solution algorithms.

Among the most representative recent studies for the minmax mTSP, Karabulut  et al. \cite{karabulut2021modeling} presented an evolution strategy approach for the mTSP with minsum and minmax objectives where they used a self-adaptive Ruin and Recreate heuristic to generate offspring and a local search to further improve the new solutions. This algorithm showed competitive results. He and Hao \cite{he2022hybrid} presented a hybrid search with neighborhood reduction (HSNR) where a dedicated strategy is used to streamline the neighborhood search by eliminating non-promising candidate solutions. Zheng et al. \cite{ZHENG2022105772} introduced an iterated two-stage heuristic algorithm (ITSHA), characterized by a special random greedy initialization procedure and an adaptive variable neighborhood search. Mahmoudinazlou and Kwon \cite{mahmoudinazlou2023hybrid} developed a hybrid genetic algorithm (HGA) with a novel crossover operator and an innovative self-adaptive random local search component. The hybrid approach makes a significant contribution to the problem-solving landscape. He and Hao \cite{he2023Memetic} proposed a memetic algorithm (MA) to solve the minmax mTSP and its generalized version with multiple depots. The algorithm includes a generalized edge assembly crossover, an efficient variable neighborhood descent, and a post-optimization phase. MA showed superior performance compared to state-of-the-art algorithms. While these approaches have been successful on a variety of benchmark instances, their performance on very challenging instances remains unsatisfactory.

In this paper, we aim to advance the state of the art in better solving the mTSP with the minmax objective by introducing a learning-guided heuristic approach. The proposed algorithm relies on the iterated local search framework \cite{LourencoMS03} and integrates ideas from adaptive large neighborhood search \cite{ropke2006adaptive,pisinger2019large}. Moreover, it exploits effective search operators and strategies from related TSP and vehicle routing problems \cite{Zhou9729095,Meng7948721}. We summarize the contributions as follows.

The proposed multi-armed bandit guided iterated local search (MILS) consists of several complementary search components. The local search procedure uses the best-improvement strategy to explore multiple neighborhoods to attain high-quality local optimal solutions. To ensure continuous diversification to visit different regions of the search space, MILS uses a probabilistic criterion to accept new local optimal solutions. To escape from the basin of attraction of deep local optima, MILS applies a learning driven perturbation (based on removal and insertion operators) to decide on the most appropriate operators to perturb the given solution. In addition, MILS benefits from an effective TSP heuristic for single route optimization.

We thoroughly evaluate the performance of the proposed algorithm on the 77 widely used benchmark instances including 41 small and medium instances (set $\mathbb{S}$) and 36 large instances (set $\mathbb{L}$). The algorithm finds record-breaking best results (new upper bounds) on 32 challenging instances considered difficult to improve, demonstrating its competitiveness and effectiveness in addressing the minmax mTSP. We also perform experiments to shed light on the rationale and understanding of the main search components of the algorithm.

Section \ref{algo} presents the proposed algorithm. Section \ref{com_results} shows an experimental comparison with state-of-the-art methods in the literature. Section \ref{analysis} presents additional experiments to analyze the main components of the algorithm, followed by conclusions and perspectives in Section \ref{conclusion}.

\section{Multi-armed bandit-driven iterated local search}\label{algo}

The proposed MILS algorithm for the minmax mTSP follows the general approach of iterated local search \cite{LourencoMS03}, and iterates three main phases: local optima exploration (Section \ref{Local optima exploration}), probabilistic solution acceptance (Section \ref{accept_section}) and local optima escaping (Section \ref{mutation_section}).

As illustrated in Algorithm \ref{algo_ils}, the algorithm starts with an initial solution built with a randomized greedy heuristic (Section \ref{initial_algo}). The solution is then improved to attain a local optimal solution during the local optima exploration phase, which uses an efficient best-improvement local search procedure with multiple neighborhoods (Section \ref{ls_solution}). The local optima exploration phase also includes a single tour improvement procedure, which is triggered under certain conditions and is based on a state-of-the-art TSP heuristic (Section \ref{post_section}). The improved local optimal solution is then submitted to the probabilistic solution acceptance criterion to decide whether to accept it as the new current solution. This is followed by the local optima escaping phase to get rid of the current local optimum trap and lead the search to a new basin of attraction. For this purpose, the local optima exploration phase perturbs the current solution by a learning-driven removal-and-insertion procedure with multi-armed bandit (MAB) to generate a new solution for the next round of the search. Finally, if the search is considered to have sufficiently explored the current search region (indicated by the condition $Iter > I_{max}$ where $Iter$ is the iteration number of the algorithm and $I_{max}$ is a parameter), the algorithm starts its next search round by creating a fresh initial solution using the randomized greedy heuristic. The algorithm stops and returns the global best solution $\varphi^{*}$ when the predetermined stopping condition is met, such as reaching a cutoff time limit or a specified number of iterations, and.

\renewcommand{\baselinestretch}{0.8}\huge\normalsize
\begin{algorithm}[]\label{algo_ils}
\begin{small}
\caption{Main framework of the MILS algorithm}
\DontPrintSemicolon 
%\KwIn{Instance $I$, parameter $I_{max}$, \textcolor[rgb]{1,0,0}{$I_{threshold}$};}
\KwIn{Instance $I$, parameter $I_{max}$;}
\KwOut{The best solution $\varphi^{*}$ found;}
\Begin{
$\varphi= GreedyRandom(I)$; \tcc*[l]{$\varphi$ represents the current solution, Section \ref{initial_algo}}
$\varphi' \gets \varphi$;\tcc*[l]{$\varphi'$ is the local optimal solution of the current round of the search}
$\varphi^{*} \gets \varphi$;\tcc*[l]{$\varphi^{*}$ is the global best solution found so far}
\textit{Iter}$\gets0$;\tcc*[l]{Iteration counter}
\While{$Stopping\ condition\ is\ not\ met$}{
	\tcc*[l]{Local optima exploration}
	$\varphi \gets $\textit{LocalSearch}($\varphi$);\tcc*[l]{Improve the current solution, Section \ref{ls_solution}}
%	\If{$f(\varphi) < f(\varphi^{*})$ and $Iter> I_{threshold}$}{
	\If{$f(\varphi) < f(\varphi^{*})$}{
		$\varphi \gets SingleTourImprove(\varphi)$;\tcc*[l]{Further improve each elite solution, Section \ref{post_section}}
	}
	\tcc*[l]{Probabilistic solution acceptance}
	<$\varphi,\varphi',\varphi^{*}$> $\gets$ $AcceptStrategy$($\varphi,\varphi',\varphi^{*}$);\tcc*[l]{Section \ref{accept_section}}
	
	\tcc*[l]{Local optima escaping}
	$D_i \gets MAB(\mathcal{D})$;$R_i \gets MAB(\mathcal{R})$;\tcc*[l]{Select removal and insertion operators with multi-armed bandit, Section \ref{mab_section}}
		$\varphi \gets Perturb(\varphi,D_i,R_i)$;\tcc*[l]{Perturb the current solution, Section \ref{des_section}}
	\tcc*[l]{Restarting}
	\eIf{$Iter > I_{max}$}{
		$\varphi= GreedyRandom(I)$;\tcc*[l]{Creat a new initial solution}
		$\varphi' \gets \varphi$;	$Iter  \gets 0$;\\
	}{
		$Iter \gets Iter + 1$;\\
	}
}
\Return{$\varphi^{*}$};\\
}
\end{small}
\end{algorithm}
\renewcommand{\baselinestretch}{1.0}\huge\normalsize

\subsection{Initial solution}\label{initial_algo}
MILS uses the following greedy randomized heuristic to build initial solutions with $m$ tours. 

\begin{enumerate}
	\item Initiate $m$ tours, where each tour starts at the depot $v_0$ and includes a randomly chosen city; 
	\item Identify the shortest tour $r$ among the $m$ tours under construction and let $v$ be the last city of tour $r$;
	\item Identify the city $u$ among the $\alpha$ nearest cities of $v$ (see Section \ref{ls_solution}) such that city $u$ causes the least increase in length for the shortest tour;
	\item Insert the city $u$ into the shortest tour after $v$;
  \item Repeat Steps (2) to (4) until all cities are assigned to routes.
\end{enumerate}
The time complexity of the greedy randomized heuristic algorithm is bounded by $\mathcal{O}(n \times \alpha)$.

\subsection{Local optima exploration} \label{Local optima exploration}

The local optima exploration phase is a key search component within the MILS algorithm. Specifically, it includes a local search with multiple neighborhoods, which are exploited with the best-improvement strategy to explore various local optimal solutions and a single tour improvement procedure with the state-of-the-art TSP heuristic EAX-TSP \cite{nagata2013powerful} to improve each individual tour. In particular, due to the time-consuming nature of EAX-TSP, the single tour improvement procedure is applied only to an elite solution $\varphi$ when it updates the global best solution $\varphi^{*}$ (i.e., $f(\varphi) < f(\varphi^{*})$). %This cautious activation of single tour improvement is mainly due to the time-consuming nature of EAX-TSP. % and helps the MILS algorithm avoid premature convergence. %, and the number of iterations exceeds a predefined threshold ($I > I_{threshold}$)

\subsubsection{Local search}\label{ls_solution}

The local search procedure uses the best-improvement strategy to exploit multiple neighborhoods generated by ten neighborhood operators (also called move operators), which are explored by the local search procedure within the framework of variable neighborhood descent \cite{mladenovic1997variable}. Note that these operators have been widely used in various routing problems.  
%Local search is a fundamental element of state-of-the-art metaheuristics applied to various routing problems \cite{liao2020integrated,che2022tabu,he2021grouping,he2022general}. The essence of local search is to start with a candidate solution and iteratively move to the better one in the immediate neighborhood of the current one. The iterative refinement process is often facilitated by the use of a variety of enriched neighborhood operators and acceleration techniques. In MILS, which is tailored to address the minmax mTSP, a notable feature is the incorporation of ten widely recognized neighborhood operators that are systematically explored to enrich the solution space. It is worth highlighting that, during the exploration, the best-improvement strategy is used to evaluate and select each candidate neighborhood solution.

Let vertex $v \in \mathcal{V}$ be a $\alpha$-nearest neighbor of city $u\in \mathcal{N}$, where $\alpha$ ($\alpha < |\mathcal{N}|$) is the granularity threshold that restricts the search to nearby vertices. Let $r(u)$ and $r(v)$ denote two tours which visit vertices $u$ and $v$, respectively. Additionally, let $x$ and $y$ represent the successors of $u$ in $r(u)$ and $v$ in $r(v)$, respectively. Then the ten neighborhood operators are defined as follows.

\begin{itemize}
\item M1: city $u$ is removed from $r(u)$ and inserted into $r(v)$ after vertex $v$.
\item M2: Two consecutive cities $u$ and $x$ are removed from $r(u)$ and inserted into $r(v)$ after vertex $v$.
\item M3: Two consecutive cities $u$ and $x$ are removed from $r(u)$ and ($x$, $u$) are placed before vertex $v$.
\item M4: Interchange the position of city $x$ and city $v$.
\item M5: Interchange two consecutive cities ($u$, $x$) and city $v$.
\item M6: Interchange two consecutive cities ($u$, $x$) and two consecutive cities ($v$, $y$).
\item M7: Interchange two consecutive cities ($x$, $u$) and two consecutive cities ($v$, $y$).
\item M8: This is the 2-opt operator, which replaces edges $(u, x)$ and $(v, y)$ by $(u, v)$ and $(x, y)$ if $r(u) = r(v)$.
\item M9: This is the 2-opt*, which replaces edges ($u,x$) and ($v,y$) with edges ($u,y$) and ($v,x$).
\item M10: This is the 2-opt*, which replaces edges ($u,x$) and ($v,y$) with edges ($u,v$) and ($x,y$).
\end{itemize}

Due to the nearest neighbor rule, the computational complexity of these operators is bounded by $\mathcal{O}(n \times \alpha)$. Unlike the two state-of-the-art algorithms, HSNR \cite{he2022hybrid} and MA \cite{he2023Memetic}, the cross-exchange operator is excluded from MILS duo to its relatively high computational cost compared to the ten move operators we used. Furthermore, inspired by the concept of \textit{don't look bits} \cite{bentley1992fast}, we have incorporated this acceleration technique to speed up our neighborhood searches by filtering out unpromising neighborhood solutions.

The local search procedure explores local optimal solutions with the best-improvement strategy by considering these neighborhoods in the listed order. In Section \ref{benefit_ls}, we also examine the first-improvement strategy, which is another popular neighborhood search strategy largely used in the context of minimizing the total traveling cost (the minsum objective) for various routing problems \cite{accorsi2021fast,vidal2022hybrid,he2022general,he2023hybrid}. Indeed, although no significant differences have been observed between these two strategies for the minsum objective, little is known regarding the behavior of these two strategies for minmax problems like the minmax mTSP. In Section \ref{benefit_ls}, we partially fill this gap by reporting experimental evidence to show that the best-improvement strategy dominates the first-improvement strategy for the minmax objective. 

\subsubsection{Single tour improvement}\label{post_section}

In addition to the aforementioned local search procedure, the local optima exploration also includes a single tour improvement procedure to further improve each new elite solution. For this, the single tour improvement procedure uses the edge assembly crossover designed for the TSP (EAX-TSP)\footnote{The code of EAX-TSP is available at: https://github.com/sugia/GA-for-TSP} to minimize each individual tour of the solution. Specifically, for each tour, its cities are submitted to EAX-TSP, which then optimizes the order of these cities and returns a new tour. After EAX-TSP is finished, its final solution is given to the local search procedure, which can possibly lead to further improvement by exchanging cities between tours.

Finally, since EAX-TSP is time consuming, we apply it only to high-quality elite solutions, which are typically reached when the algorithm has searched long enough. Specifically, the single tour improvement procedure is only triggered when the underlying solution ($\varphi$) updates the global best solution $\varphi^{*}$ ($f(\varphi) < f(\varphi^{*})$) and the number of iterations reaches a threshold $I_{threshold}$ (empirically set to $I_{threshold}=1000$).

\subsection{Probabilistic acceptance and stopping criterion}\label{accept_section}

Let $\varphi$ be the solution from the local optima exploration phase, and let $\varphi'$ be the current local optimal solution. We use a probabilistic acceptance criterion to decide whether $\varphi$ is accepted to update $\varphi'$. The decision to incorporate $\varphi$ into the subsequent trajectory of the search process follows the acceptance criterion of simulated annealing \cite{accorsi2021fast,maximo2021hybrid}. As shown in Algorithm \ref{algo:sa}, within each iteration of MILS, if the solution $\varphi$ is better than the global best solution $\varphi^{*}$ ($f(\varphi) < f(\varphi^{*})$), $\varphi$ is accepted. Consequently, both $\varphi^{*}$ and the current local optimal solution $\varphi'$ undergo corresponding updates. However, if $f(\varphi^{*}) < f(\varphi) < f(\varphi')$, only the current local optimal solution $\varphi'$ receives an update. In the event that $f(\varphi) > f(\varphi')$, the acceptance of the solution $\varphi$ depends on a probability determined by $e^{-\Delta f/T}$, where $\Delta f = f(\varphi) - f(\varphi')$ and $T$ is the temperature parameter, which is tuned according to Eqs. \ref{T} and \ref{cooling}. The initial temperature $T_0$ is determined by Eq. \ref{initial_tem}, where $\varphi_{init}$ is the given initial solution and $w$ is a parameter set to 0.35 following \cite{ropke2006adaptive}. The final temperature $T_f$ over $I_{max}$ iterations are set to $0.0001$ empirically.

\renewcommand{\baselinestretch}{0.8}\huge\normalsize
\begin{algorithm}[]\label{algo:sa}
\begin{small}
\caption{Acceptance criterion}
\DontPrintSemicolon 
\KwIn{Solution $\varphi$ from the local optima exploration procedure, current local optimal solution $\varphi'$, global best solution $\varphi^{*}$;}
\KwOut{<$\varphi$, $\varphi'$, $\varphi^{*}$>;}
\Begin{
	$\Delta f\gets f(\varphi) - f(\varphi')$;\\
	\uIf{$f(\varphi) < f(\varphi^{*})$}{
		$\varphi^{*} \gets \varphi$;	$\varphi' \gets \varphi$;\\
	}
	\uElseIf{$f(\varphi) < f(\varphi')$}{
		$\varphi' \gets \varphi$;\\
	}
	\uElseIf{$e^{-\Delta f/T} > random(0,1)$}{
		$\varphi' \gets \varphi$;\tcc*[l]{Accept $\varphi$ as new current local optimal solution}
	}
	\Else{
	$\varphi \gets \varphi'$;\tcc*[l]{Refuse $\varphi$ as new current local optimal solution}
	}
\Return{<$\varphi$,$\varphi'$,$\varphi^{*}$>};\\
}
\end{small}
\end{algorithm}
\renewcommand{\baselinestretch}{1.0}\huge\normalsize

\begin{equation}\label{T}
T_{Iter+1} = cT_{Iter}
\end{equation}
\begin{equation}\label{cooling}
c = (\frac{T_{f}}{T_0})^{I_{max}}
\end{equation}
\begin{equation}\label{initial_tem}
e^{-(w\cdot f(\varphi_{init}))/T_0} = p_{accept}
\end{equation}

\subsection{Local optima escaping}\label{mutation_section}

When the local optima exploration procedure stagnates, the search is trapped in a deep local optimum. To escape from the current basin of attraction and explore alternative neighboring basins of attraction, the local optima escaping procedure is triggered to perturb the current solution. In the proposed algorithm, we use a set of removal operators ($\mathcal{D}$) and a set of insertion operators ($\mathcal{R}$) to perturb the solution $\varphi$ by selectively removing and then reinserting some cities. For this, a pair of removal and insertion operators is chosen from the sets $\mathcal{D}$ and $\mathcal{R}$, respectively. The selection can be performed randomly or using the roulette-wheel strategy as in \cite{ropke2006adaptive}. Here, we adopt the multi-armed bandit algorithm to intelligently identify the appropriate removal and insertion operators.

\subsubsection{Removal and insertion operators}\label{des_section}

We use five removal operators: \textit{Shaw removal}, \textit{Random removal}, \textit{Cross removal}, \textit{Worst removal} and \textit{Information removal}. Each removal operator removes $\lfloor l \times n\rfloor$ cities and keep them in set $\mathcal{S}$, where $l$ ($0 < l < 1$) is a parameter called the length of perturbation \cite{benlic2013breakout}.

\begin{enumerate}
\item \textit{Shaw removal.} The operator removes $\lfloor l \times n\rfloor$ cities based on a given criterion, such as geographic location \cite{shaw1998using}. Specifically, a random city $v_i$ is used to initialize an empty set $\mathcal{S}$. Then the similarity between the city $v_i$ and the remaining cities in the set $\mathcal{N}\setminus\mathcal{S}$ is computed. The similarities are then used to sort the remaining cities in ascending order in the list $\mathcal{L}$. To introduce randomness into the selection of cities to remove, a parameter denoted as $\gamma$ (where $\gamma \geq 1$) is used. Then, given a random number $y$ from $random(0,1)$, $\lfloor y^{\gamma}*|\mathcal{L}| \rfloor$th city ($u_i$) in the list $\mathcal{L}$ is selected and removed from the solution and added in the set $\mathcal{S}$ ($\mathcal{S}\gets\mathcal{S}\cup\{u_i\}$). The newly selected city $u_i$ is used to select the next city according to the above procedure. The operation continues until $\lfloor l \times n\rfloor$ cities are successfully identified and selected for removal.

\item \textit{Random removal.} The operator randomly selects $\lfloor l\times n\rfloor$ cities, removes them from the solution $\varphi$ and inserts them into an initially empty set $\mathcal{S}$.

\item \textit{Cross removal.} The operator focuses on removing cities that are close to each other, primarily based on their geographic location. In fact, cities with close coordinates, even if assigned to different tours, have an increased probability of being reassigned. The reassignment serves as a means of perturbing the solution, making it easier for the algorithm to escape from local optima. Specifically, for each city $v_i$, the operator incrementally counts the number of neighboring cities in its vicinity, denoted by $\alpha$, that are visited by alternative tours. Consequently, if a majority of a city's neighbors are actually served by different tours, it becomes more likely that this city will be moved to increase the chances of escaping local optima trap. Thus, cities are stored and sorted in $\mathcal{L}$ according to the number of neighbors visited by alternative tours. Similar to the \textit{Shaw removal} operator, we also use a parameter $\gamma$ to introduce randomness into the selection of cities.

\item \textit{Worst removal.} The objective of the operator is to identify and remove cities that contribute to a significant increase in the total traveling cost along the tour. The operator first calculates for each city the potential reduction in traveling costs that would result from removing it from the solution. These computed values are then used to sort the cities in descending order $\mathcal{L}$, based on the magnitude of these reductions. Similar to the \textit{Shaw removal} operator, the selection of cities to remove involves an element of randomization, moderated by a deterministic parameter $\gamma$. The parameter controls the degree of randomization in the selection of the 'worst' city to remove from the solution.

\item \textit{Information Removal.} It is a well-established observation that high-quality solutions to routing problems often exhibit structural proximity to the optimal solution, sharing a significant number of common edges with them \cite{he2022general}. In particularly, these solutions tend to exhibit recurring sequences of consecutive visits, commonly referred to as \textit{patterns}, which are typically not actively explored during the standard local search procedure \cite{arnold2021pils}. Consequently, if some cities are frequently involved in neighborhood operators, this tendency may be advantageous in allowing MILS to escape local optima. To take advantage of this insight, a statistical analysis is performed during the local search procedure to track the frequency with which each city is involved in the neighborhood operators. Then, all cities are ordered in $\mathcal{L}$ in descending order based on their involvement frequency. Similar to the \textit{Shaw removal} operator, the selection of cities for removal introduces an element of randomness, moderated by a parameter $\gamma$. The procedure terminates when a specified number of $\lfloor l\times n\rfloor$ cities have been successfully removed.
\end{enumerate}

After the application of the removal operators, the removed cities (retained in the set $\mathcal{S}$) are reinserted into the solution by using three insertion operators: \textit{Greedy insert}, \textit{Greedy insert with blink}, and \textit{Regret insert}. 

\begin{enumerate}
\item \textit{Greedy insertion.} For each city $v_i \in \mathcal{S}$, a city $u_j \in \mathcal{N} \setminus \mathcal{S}$, which comes from its $\alpha$-nearest neighborhood and minimizes the traveling cost of the tour after insertion, is chosen. Then, city $v_i$ is inserted after city $u_j$. The process continues until all cities in the set $\mathcal{S}$ have been successfully inserted into the solution.

\item \textit{Greedy insertion with blink.} The operator, originally proposed in \cite{christiaens2020slack} for the vehicle routing problem, introduces a controlled element of randomness into the insertion process. For each city $v_i \in \mathcal{S}$, a city $u_j \in \mathcal{N} \setminus \mathcal{S}$ from its $\alpha$-nearest neighborhood is selected for insertion with a probability of $1-\beta$. If the probability is not met, the insertion position is skipped. The process continues until all cities in the set $\mathcal{S}$ have been successfully inserted into the solution. Following \cite{christiaens2020slack}, we set the parameter $\beta = 0.01$ to control the amount of randomness introduced during the insertion procedure.

\item \textit{Regret insertion.} In the operator, the selection process at each iteration focuses on identifying the removed cities that would cause the most regret if they were not inserted at their optimal position during the current iteration. For each city $v_i \in \mathcal{S}$, we first compute the cost, denoted as $\Delta f^{1}_{i}$, associated with inserting $v_i$ at its best possible position within the incumbent solution. Additionally, we compute $\Delta f^{q}_{i}$ for $q \in \{2,\cdots,k\}$, representing the cost of inserting $v_i$ at its $q^{th}$ best position. Next, for each city $v_i \in \mathcal{S}$, we compute a regret value $reg_i = \sum^{k}_{q=2}(\Delta f^{q}_i-\Delta f^{1}_i)$. Finally, we select the position with the highest regret value, and insert the city $v_i$ at that location. Following \cite{ropke2006adaptive}, the parameter $k$ is set to 3. 
\end{enumerate}

\subsubsection{Multi-armed bandit}\label{mab_section}

Following the idea of \cite{lagos2023multi} where multi-armed bandit was used to select lower-level heuristics within the framework of hyper-heuristics, we explore the use of the multi-armed bandit algorithm for selecting removal and insertion operators to perturb solutions. 

For the removal operators, each operator $D_i \in \mathcal{D}$ is associated with three critical attributes: a weight $\omega_{D_i}$, a score $\pi_{D_i}$, and a counter $\theta_{D_i}$ that records the number of times the operator $D_i$ has been applied. Initially, all operators are given the same weight, with the constraint that the sum of these weights over all operators in $\mathcal{D}$ equals 1. We divide the algorithm execution into segments of fixed duration (100 iterations per segment in our case), with constant operator weights maintained throughout each segment. Before starting a new segment, the scores ($\pi_{D_i}$) of all operators are reset to zero. During each iteration within a segment, the score of the selected operator $D_i$ is updated by $\pi_{D_i} \gets \pi_{D_i} + \delta_u$, where $\delta_u$ ($u \in \{1,2,3\}$) contains three different cases that govern the score change $\pi_{D_i}$. Specifically, the score $\pi_{D_i}$ receives an increase of $\delta_1$ when the global best solution $\varphi^{*}$ undergoes an update ($f(\varphi) < f(\varphi^{*})$). Similarly, an increase of $\delta_2$ is received when the local optimum solution $\varphi'$ is improved ($f(\varphi) < f(\varphi')$), and a score increment of $\delta_3$ is applied when solution $\varphi$ is accepted according to the probabilistic solution acceptance criterion, even if it does not represent an overall improvement. In our case, we used $\delta_u = 3,5,10$, for $u=1,2,3$. At the end of each segment, we compute new weights for the operators based on the accumulated scores according to Eqs. \ref{ccc} and \ref{ccc3}, where $\omega_{D_i,j}$ is the weight of operator $D_i$ in last segment $j$ and $\lambda$ is the reaction factor which controls how quickly the weight adjustment reacts to changes.

Then, MAB selects operators as follows. If $r_n< \epsilon$, where $r_n$ is a random number between (0,1) and $\epsilon$ is the parameter of $\epsilon$-greedy, the operator associated with the maximum weight is selected; otherwise, an operator is selected at random. 

\begin{equation}\label{ccc}
\pi_{D_i} = \frac{\pi_{D_i}}{\theta_{D_i}}
\end{equation} 
\begin{equation}\label{ccc3}
\omega_{D_i,j+1} = (1-\lambda)\omega_{D_i,j} + \lambda {\pi}_{D_i}
\end{equation}

\section{Experimental Evaluation and Comparisons}\label{com_results}

The purpose of this section is to evaluate the performance of MILS through experiments and compare it with state-of-the-art algorithms.

\subsection{Benchmark instances}\label{instances}

We conducted experiments using two benchmark sets that are commonly tested in the literature. The instances (77 in total) and the best solutions obtained by MILS are available online\footnote{https://github.com/pengfeihe-angers/mils.git}.

\begin{itemize}
\item Set $\mathbb{S}$: The set contains 41 small and medium instances with 51 to 1173 cities and 3 to 30 tours. These instances have been widely used to evaluate minmax mTSP algorithms \cite{he2022hybrid,he2023Memetic,venkatesh2015two,wang2017memetic,ZHENG2022105772,mahmoudinazlou2023hybrid}. Among these 41 instances, optimal solutions are known for 17 instances including the two instances with 30 tours. Given that the instances of this set have been studied intensively, it is challenging to improve the best-known results for the instances with unknown optima.
\item Set $\mathbb{L}$: The set has 36 large instances with 1357 to 5915 cities and 3 to 20 tours, which was introduced in \cite{he2022hybrid} and further tested in \cite{he2023Memetic}. Among these 36 instances, optimal solutions are known for 5 instances including the two instances with 30 tours. The $\mathbb{L}$ instances are known to be more challenging than those of the set $\mathbb{S}$. % since it requires that algorithms should converge quickly and guarantee solutions' quality. 
\end{itemize}

\subsection{Experimental protocol and reference algorithms}\label{exper_prococol}

\textbf{Parameter setting.} The MILS algorithm has five parameters: the maximum number of iterations per restart $I_{max}$, the accept probability $p_{accept}$, the nearest neighbor granularity threshold $\alpha$, the perturbation length $l$, the $\epsilon$-greedy parameter $\epsilon$. In order to calibrate these parameters, the automatic parameter tuning package Irace \cite{lopez2016irace} was used. The tuning was performed on 8 instances with 150-2392 cities and the tuning budget was set to be 2000 runs. The candidate and final values for the parameters are shown in Table \ref{table_parameter}. The values recommended by Irace were used throughout our experiments and can be considered as the default parameter setting of the MILS algorithm.

\begin{table}
\caption{Parameter tuning results.} \label{table_parameter}
\renewcommand{\baselinestretch}{}\huge\normalsize
\begin{tiny}
\begin{tabular}{p{0.6cm}p{0.5cm}p{1.5cm}p{3.5cm}p{0.8cm}}
\hline
Parameter & Section & Description & Considered values & Final values\\
\hline
$I_{max}$						&\ref{accept_section}  & maximum iterations per restart &$\{10000,20000,30000,40000,50000\}$  &40000\\
$p_{accept}$						& \ref{accept_section} & accept probability & $\{0.5,0.6,0.7,0.8,0.9\}$ &0.7\\
$\alpha$							&\ref{ls_solution}          & granularity threshold & $\{5,10,15,20,25,30\}$ &10 \\
$l$								&\ref{mutation_section} & perturbation length& $\{0,0.05,0.1,0.15,0.2,0.25,0.3\}$  &0.15\\
$\epsilon$									&\ref{mutation_section}&$\epsilon$-greedy parameter & $\{0.005,0.01,0.015,0.02,0.025\}$ &0.01\\
\hline
\end{tabular}
\end{tiny}
\end{table}

\textbf{Reference algorithms.} For our comparative experiments, we used the following four recent state-of-the-art algorithms, which hold most of the current best-known solutions for the minmax mTSP benchmark instances. 

\begin{itemize}

\item BKS. This indicates the best-known solutions (current best upper bounds) that are reported by the state-of-the-art algorithms \cite{he2022hybrid,ZHENG2022105772,mahmoudinazlou2023hybrid,he2023Memetic,karabulut2021modeling}.

\item HSNR \cite{he2022hybrid} (2022). This hybrid search algorithm with neighborhood reduction was coded in C++ and executed on a computer with a Xeon E5-2670 CPU at 2.5 GHz and 6 GB RAM. 

\item ITSHA \cite{ZHENG2022105772} (2022). This iterated two-stage heuristic algorithm) was implemented in C++, and we ran the source code on our computer (Xeon E5-2670 CPU at 2.5 GHz and 6 GB RAM).

\item HGA \cite{mahmoudinazlou2023hybrid} (2023). This hybrid genetic algorithm was coded in Julia and executed on a computer with an Apple M1 CPU and 16 GB RAM. According to Passmark\footnote{https://www.cpubenchmark.net/singleThread.html\#}, the M1 CPU is significantly faster than the Xeon E5-2670 processor used in this study, in terms of the single thread performance. Therefore, given the same running time, the HGA algorithm has a more favorable computational budget.

\item MA \cite{he2023Memetic} (2023). This mementic algorithm was implemented in C++ and executed on a computer with a 2.5 GHz Xeon E5-2670 processor and 8 GB of RAM. 

\end{itemize}

\textbf{Experimental setting and stopping criterion.} 
We implemented the MILS algorithm in C++ and compiled the program using the g++ compiler with the -O3 option\footnote{The code of MILS will be available at: https://github.com/pengfeihe-angers/mils.git}. All experiments were conducted on a 2.5 GHz Intel Xeon E-2670 processor with 6 GB RAM running Linux with a single thread. Given the stochastic nature of the algorithm, MILS was run 20 times on each instance with different random seeds. The algorithm terminates when it reaches the cutoff limit of $(n/100)\times 4$ minutes (this is the same stopping condition used in the reference algorithms \cite{he2023Memetic,he2022hybrid,ZHENG2022105772}).

\subsection{Computational results and comparisons}\label{comparison1}

To compare the MILS algorithm with the reference algorithms, we provide a brief summary of the results in Table \ref{sumResults} and detailed results in the appendix. We show the number of instances where our MILS algorithm obtains a better (\#Wins), equal (\#Tiers), or worse (\#Losses) result compared to each reference, including the BKS. To verify whether there is a statistically significant difference between MILS and each reference algorithm, the Wilcoxon signed-rank test is used at a 0.05 confidence level, where a \textit{p-value} less than 0.05 indicates a statistically significant difference.

%\begin{figure}[H]
%\centering
%\includegraphics[width=3.5in]{./figure/summary.pdf}
%\centering
%\caption{Performance chart of MILS on different groups.}
%\label{perfor_new}
%\end{figure}

First, regarding the BKS, Table \ref{sumResults} shows that MILS updates 32 current best-known results (new upper bounds) out of the 77 instances (41.6\%) and matches 35 other BKS values (45.5\%). The improved best-known results concerns 7 $\mathbb{S}$ instances and 25 $\mathbb{L}$ instances. These results are remarkable because the $\mathbb{S}$ instances have been studied intensively in the literature and the $\mathbb{S}$ instances are known to be challenging for the existing algorithms.

Now, if we examine the results in terms of the number of tours, we can make the following comments. For $m\in \{3,5\}$, MILS performs remarkably well compared to the BKS values, and the \textit{p-values} (<0.05) clearly indicate that the differences are statistically significant. For $m\in\{10,20\}$, MILS competes well with the BKS results too, as it still yields several new bounds. However, the \textit{p-values} ($>0.05$) indicate that the differences are marginal. As shown in Tables \ref{tableA1} and \ref{tableA2} of the appendix, for small instances ($n<532$), MILS consistently gives excellent results when $m\in \{3,5\}$, without losing a single case. For medium and large instances, our algorithm also improves many BKS results (improvement gap up to $9.03\%$), indicating that there is still room for improvement for these large instances.

\begin{table}
\caption{Summary of comparative results between MILS and the reference algorithms on the 77 benchmark instances.}\label{sumResults}
\begin{tiny}
\begin{center}
\begin{tabular}{p{0.9cm}p{0.4cm}p{0.3cm}p{0.3cm}p{0.3cm}p{0.7cm}p{0.01cm}p{0.3cm}p{0.3cm}p{0.3cm}p{0.7cm}}
\hline
\multirow{2}{*}{Pair   algorithms} & \multicolumn{1}{c}{\multirow{2}{*}{Groups}} & \multicolumn{4}{c}{Best}                       &  & \multicolumn{4}{c}{Avg.}                       \\
\cline{3-6}\cline{8-11}
                                   & \multicolumn{1}{c}{}                        & \#Wins & \#Tiers & \#Losses & \textit{p-value} &  & \#Wins & \#Tiers & \#Losses & \textit{p-value} \\
\hline
\multirow{4}{*}{MILS   vs. BKS} & \textit{m=3}  & 10 & 7  & 2 & 2.10E-02 &  & -  & -  & - & -        \\
                                & \textit{m=5}  & 11 & 5  & 3 & 5.25E-03 &  & -  & -  & - & -        \\
                                & \textit{m=10} & 7  & 7  & 5 & 1.10E-01 &  & -  & -  & - & -        \\
                                & \textit{m=20} & 4  & 14 & 0 & 1.25E-01 &  & -  & -  & - & -        \\
                                & \textit{m=30} & 0  & 2 & 0 & 0.00E+00 &  & -  & -  & - & -        \\                                
\hline
\multirow{4}{*}{MILS vs. HSNR}  & \textit{m=3}  & 16 & 3  & 0 & 4.38E-04 &  & 19 & 0  & 0 & 1.32E-04 \\
                                & \textit{m=5}  & 16 & 1  & 2 & 6.29E-04 &  & 17 & 1  & 1 & 7.38E-04 \\
                                & \textit{m=10} & 10 & 7  & 2 & 9.28E-03 &  & 10 & 6  & 4 & 3.40E-02 \\
                                & \textit{m=20} & 4  & 14 & 0 & 1.25E-01 &  & 4  & 14 & 0 & 1.25E-01 \\
                                & \textit{m=30} & 0  & 2 & 0 & 0.00E+00 &  & 0  & 2 & 0 & 0.00E+00   \\                                                                
\hline
\multirow{4}{*}{MILS vs. ITSHA} & \textit{m=3}  & 16 & 3  & 0 & 4.38E-04 &  & 17 & 2  & 0 & 2.93E-04 \\
                                & \textit{m=5}  & 17 & 2  & 0 & 2.93E-04 &  & 16 & 1  & 2 & 1.18E-03 \\
                                & \textit{m=10} & 11 & 7  & 1 & 3.42E-03 &  & 12 & 3  & 5 & 1.48E-02 \\
                                & \textit{m=20} & 7  & 11 & 0 & 1.56E-02 &  & 13 & 5  & 0 & 2.44E-04 \\
                                & \textit{m=30} & 0  & 2 & 0 & 0.00E+00 &  & 0  & 2 & 0 & 0.00E+00   \\
\hline
\multirow{4}{*}{MILS vs. HGA}   & \textit{m=3}  & 6  & 4  & 0 & 3.13E-02 &  & 6  & 2  & 2 & 1.09E-01 \\
                                & \textit{m=5}  & 7  & 3  & 0 & 1.56E-02 &  & 6  & 2  & 2 & 7.81E-02 \\
                                & \textit{m=10} & 1  & 6  & 3 & 6.25E-01 &  & 1  & 3  & 6 & 2.19E-01 \\
                                & \textit{m=20} & 0  & 9  & 0 & 0.00E+00 &  & 2  & 7  & 0 & 5.00E-01 \\
                                & \textit{m=30} & 0  & 2 & 0 & 0.00E+00 &  & 0  & 2 & 0 & 0.00E+00   \\                                
\hline
\multirow{4}{*}{MILS vs. MA}    & \textit{m=3}  & 10 & 7  & 2 & 2.10E-02 &  & 11 & 3  & 5 & 4.94E-02 \\
                                & \textit{m=5}  & 11 & 5  & 3 & 5.25E-03 &  & 10 & 2  & 7 & 1.02E-01 \\
                                & \textit{m=10} & 7  & 7  & 5 & 6.40E-02 &  & 9  & 4  & 7 & 8.79E-02 \\
                                & \textit{m=20} & 7  & 11 & 0 & 1.56E-02 &  & 11 & 7  & 0 & 9.77E-04 \\
                                & \textit{m=30} & 0  & 2 & 0 & 0.00E+00 &  & 0  & 2 & 0 & 0.00E+00   \\                                
\hline
\end{tabular}
\end{center}
\end{tiny}
\end{table}

\begin{table}
\caption{Summary improvement information of MILS on different groups.}\label{perfor_new}
\begin{tiny}
\begin{center}
\begin{tabular}{p{2.8cm}p{0.5cm}p{0.5cm}p{0.5cm}p{0.5cm}}
\hline
Index &m=3 &m=5 & m=10  &m=20\\
\hline
Average gap (\%) to BKS   & -0.26&	-0.87 &	-1.39 &	-1.28\\
Number of improved instances & 10 & 11 & 7 &4\\
\hline
\end{tabular}
\end{center}
\end{tiny}
\end{table}

Compared to the reference algorithms, MILS significantly outperforms HSNR \cite{he2022hybrid} and ITSHA \cite{ZHENG2022105772} (\textit{p-values} $<0.05$) on the instances with $m\in\{3,5,10\}$. Furthermore, MILS achieves a remarkable performance compared to the two latest algorithms HGA \cite{mahmoudinazlou2023hybrid} and MA \cite{he2023Memetic}, losing only in a few of cases. To further verify the performance of MILS on instances with different tours, Table \ref{perfor_new} shows the average gap of MILS to the BKS results. We can observe that MILS can improve more instances with $m\in\{3,5\}$, but the improvement is marginal. On the contrary, although MILS can improve fewer instances with $m\in\{10,20\}$, the improvements are more significant. In conclusion, we can say that MILS is particularly competitive when solving instances with relatively few tours. Note that HSNR \cite{he2022hybrid} and MA \cite{he2023Memetic} can achieve remarkable results on instances with $m\in\{10\}$, making MILS a suitable algorithm for challenging instances with $m\in\{3,5,20\}$. 

\section{Additional experiments}\label{analysis}

This section presents experiments to gain additional insights into the influences of the components of the MILS algorithm: the local search procedure, the MAB algorithm. All experiments are based on the two sets of 77 instances. In addition, we also investigate the long-term convergence behavior of the algorithm under a relaxed stopping condition. To ensure a fair comparison, all variants of MILS were run on the same machine with the same parameters as MILS, and each variant was run 20 times per instance.
%A time-to-target analysis to investigate the computational efficiency of the compared algorithms.

\subsection{Rational behind the local search procedure}\label{benefit_ls}

\begin{figure}[]
\centering
\includegraphics[width=3.2in]{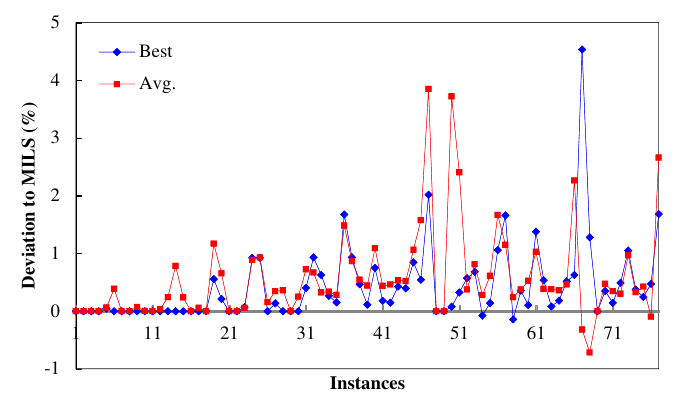}
\centering
\caption{Comparative results of MILS with variant MILS$_0$ on the 77 instances.}
\label{perfor_ls}
\end{figure}

The local search procedure of MILS uses the best-improvement strategy to explore each of the used neighborhood (see Section \ref{ls_solution}). One question is that how MILS performs when the best-improvement strategy is replaced by the first-improvement strategy, which is much more time efficient and very popular for routing problems with the minsum objective. To perform a rigorous evaluation of these two strategies, we compared a variant of MILS (called MILS$_0$) where the first-improvement strategy is used in the local search procedure. The results of the comparison are summarized in Fig. \ref{perfor_ls} and Table \ref{compar_ls}.

\begin{table}[]
\caption{Summary of comparative results between MILS and MILS$_0$.}\label{compar_ls}
\begin{tiny}
\begin{center}
\begin{tabular}{p{1.4cm}p{0.2cm}p{0.3cm}p{0.3cm}p{0.7cm}p{0.01cm}p{0.2cm}p{0.3cm}p{0.3cm}p{0.7cm}}
\hline
\multicolumn{1}{c}{\multirow{2}{*}{Pair algorithms}} & \multicolumn{4}{c}{Best}                       &  & \multicolumn{4}{c}{Avg.}                       \\
\cline{2-5}\cline{7-10}
\multicolumn{1}{c}{}     & \#Wins & \#Tiers & \#Losses & \textit{p-value} &  & \#Wins & \#Tiers & \#Losses & \textit{p-value} \\
\hline
MILS vs MILS$_0$        & 49     & 26      & 2       & 1.08E-08       &  & 60     & 14      & 3       & 2.18E-09        \\
\hline
\end{tabular}
\end{center}
\end{tiny}
\end{table}

According to Table \ref{compar_ls}, the best-improvement strategy dominates the first-improvement strategy, and it is critical to ensure the high performance of the MILS algorithm. In fact, using the first-improvement strategy significantly worsens the results of both the best and average results. Moreover, as shown in Fig. \ref{perfor_ls}, the differences become even more significant as the size of the instances increases. These results demonstrate the positive contributions of the best-improvement strategy.  

\begin{table}
\caption{Summary of average results between two improvement strategies.}\label{compar_first_best}
\begin{tiny}
\begin{center}
\begin{tabular}{p{1.8cm}p{2.0cm}p{2.0cm}}
\hline
\multicolumn{1}{c}{\multirow{2}{*}{Instances}}  & \multicolumn{2}{c}{Deviation from the original solution(\%)} \\
\cline{2-3}
\multicolumn{1}{c}{}     &  Best                          & First                        \\
\hline
att532-3  & 18.228 & 16.406 \\
pcb1173-3    & 15.565 & 14.467 \\
pr2392-5  &  15.416 & 14.331 \\
fl3795-3  &  8.575  & 8.480        \\
\hline
\end{tabular}
\end{center}
\end{tiny}
\end{table}

To gain a deeper understanding of why the best-improvement strategy is better than the first-improvement strategy, we study the local optimal solutions that can be reached by both strategies after the same number of neighborhood moves, starting from the same initial solution. Let $\varphi$ be a given initial solution. Let $\varphi_1$ and $\varphi_2$ denote the solutions refined by the local search procedure with the best-improvement and first-improvement strategies, respectively. We used four representative instances (att532-3, pcb1173-3, pr2392-5, and fl3795-3) of different sizes for this study. For each instance, we ran the local search procedure for a maximum of 10000 neighborhood moves and retained 1000 random local optimal solutions for analysis. As shown in Table \ref{compar_first_best}, for these instances, the average deviation from the original solution with the best-improvement strategy is larger than with the first-improvement strategy, which means that the best-improvement strategy allows the algorithm to go further in its search to attain high-quality solutions that cannot be reached with the first-improvement strategy. These results underscore the effectiveness of the best-improvement strategy in achieving superior results within a given number of iterations, making it the first choice for solving min-max objective routing problems.

\subsection{Rationale behind the MAB algorithm}\label{benefit_mab}

\begin{figure*}[htbp]
\centering
\subfigure[Best]{
\begin{minipage}[t]{0.5\linewidth}
\centering
\includegraphics[width=2.5in]{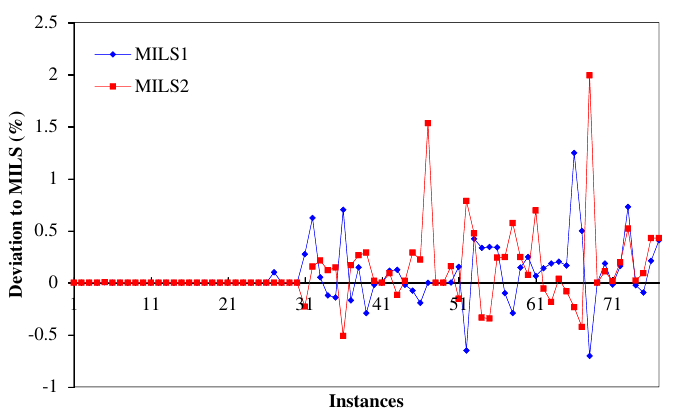}
\end{minipage}%
}%
\subfigure[Average]{
\begin{minipage}[t]{0.5\linewidth}
\centering
\includegraphics[width=2.5in]{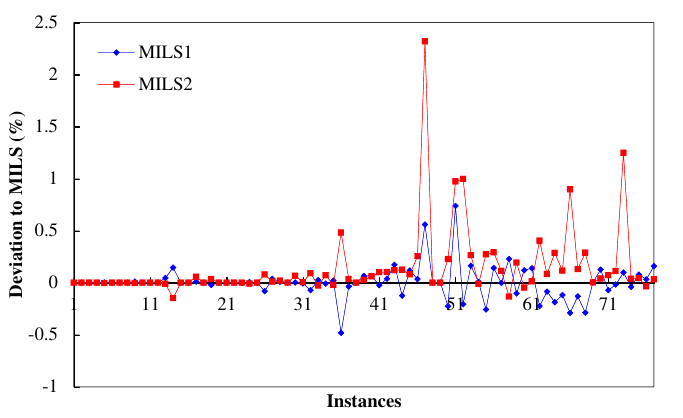}
\end{minipage}%
}%
\centering
\caption{Comparative results of MILS with two variants MILS$_1$ (with roulette-wheel strategy) and MILS$_2$ (with random strategy) on the 77 instances.}
\label{mabCompare}
\end{figure*}

To evaluate the benefit of the MAB algorithm for escaping local optima, two MILS variants (denoted by MILS$_1$ and MILS$_2$) were created where the MAB algorithm is replaced by the roulette-wheel strategy and the random strategy in MILS$_1$ and MILS$_2$, respectively. Table \ref{compar_mab} summarizes the comparative results of the three algorithms, and Fig. \ref{mabCompare} plots the best/average gap between these variants and MILS on all instances. The X-axis is the instance label, while the Y-axis is the deviation from the MILS results (\%).

\begin{table}[h]
\caption{Summary of comparative results of MILS with two variants.}\label{compar_mab}
\begin{tiny}
\begin{center}
\begin{tabular}{p{1.5cm}p{0.2cm}p{0.3cm}p{0.3cm}p{0.7cm}p{0.01cm}p{0.2cm}p{0.3cm}p{0.3cm}p{0.7cm}}
\hline
\multicolumn{1}{c}{\multirow{2}{*}{Pair algorithms}} & \multicolumn{4}{c}{Best}                       &  & \multicolumn{4}{c}{Avg.}                       \\
\cline{2-5}\cline{7-10}
\multicolumn{1}{c}{}     & \#Wins & \#Tiers & \#Losses & \textit{p-value} &  & \#Wins & \#Tiers & \#Losses & \textit{p-value} \\
\hline
MILS vs MILS$_1$        & 29     & 32      & 16       & 8.64E-03
         &  & 32     & 19      & 26       & 6.15E-01      \\
MILS vs MILS$_2$       & 33     & 33      & 11       & 5.48E-03       &  & 47     & 19      & 11       & 1.38E-06       \\
\hline
\end{tabular}
\end{center}
\end{tiny}
\end{table}

Table \ref{compar_mab} clearly shows that MILS outperforms MILS$_1$ in terms of best and average results, which is confirmed by \textit{p-values} $\ll 0.05$. As shown in Fig. \ref{mabCompare}, there are no significant differences between MILS and MILS$_1$ when solving small and medium instances ($n<532$). On the contrary, MILS$_1$ becomes worse when solving large instances, in terms of best and average results. As to the MILS$_2$ variant, Table \ref{compar_mab} shows that it performs the worst. Thus, we can conclude that the MAB algorithm is a better strategy for operator selection than the roulette-wheel strategy, and the roulette strategy is better than the random strategy. %Our experimental results provide evidences that the MAB algorithm is a crucial component of the search framework of the MILS. 

\subsection{Convergence analysis of the MILS algorithm}\label{anal_mils}

In Section \ref{comparison1}, the stopping condition of MILS was set to the maximum running time of $(n/100)\times 4$ minutes in accordance with the reference algorithms. To investigate the convergence behavior of the MILS algorithm for a long run (denoted by MILS$_{L}$), the stopping condition is set to a relaxed running time of $(n/100)\times 8$ minutes.

\begin{table}[]
\caption{Summary of comparative results between MILS and MILS$_{L}$.}\label{compar_converge}
\begin{tiny}
\begin{center}
\begin{tabular}{p{1.5cm}p{0.2cm}p{0.3cm}p{0.3cm}p{0.7cm}p{0.01cm}p{0.2cm}p{0.3cm}p{0.3cm}p{0.7cm}}
\hline
\multicolumn{1}{c}{\multirow{2}{*}{Pair algorithms}} & \multicolumn{4}{c}{Best}                       &  & \multicolumn{4}{c}{Avg.}                       \\
\cline{2-5}\cline{7-10}
\multicolumn{1}{c}{}     & \#Wins & \#Tiers & \#Losses & \textit{p-value} &  & \#Wins & \#Tiers & \#Losses & \textit{p-value} \\
\hline
MILS$_{L}$  vs MILS        & 23     & 54      & 0       & 2.95E-04       &  & 56     & 21     & 0       & 7.55E-11        \\
\hline
\end{tabular}
\end{center}
\end{tiny}
\end{table} 
 
\begin{figure}[]
\centering
\includegraphics[width=3.2in]{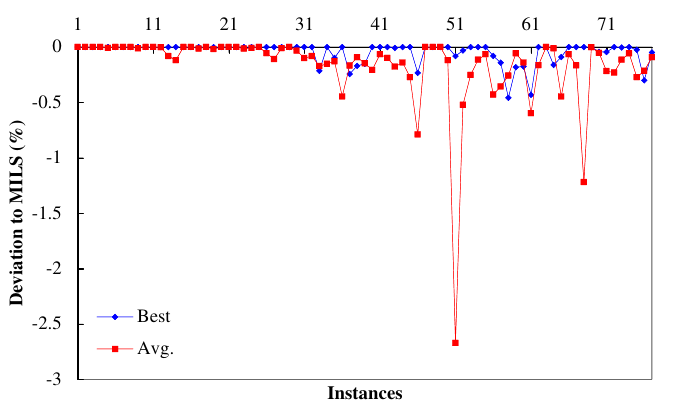}
\centering
\caption{Comparative results of MILS with variant MILS$_L$ on the 77 instances.}
\label{perfor_long}
\end{figure}

From Table \ref{compar_converge}, one observes that with a large runtime budget, MILS is able to further significantly improve its results obtained with the default cutoff time ($n/100 \times$ 4 minutes) both in terms of best and average results (\textit{p-values} $\ll 0.05$).  Specifically, the best results can be even improved as much as 0.46\%, while the average results can be improved by 2.67\%. As shown in Fig. \ref{perfor_long}, the improvements only concern medium and large instances ($n\geq 532$), which probably indicates that the results of MILS with the default cutoff time ($n/100 \times$ 4 minutes) for the small instances, if not optimal, could be very close to the optimal results. This experiment shows that the MILS algorithm has the highly desirable ability to find even better results with a longer cutoff limit for large and difficult instances.

\section{Conclusions}\label{conclusion}

%\textcolor[rgb]{1,0,0}{[I'll work on this part later]}

We introduce the first multi-armed bendit algorithm driven iterated local search for solving the minmax mTSP. The algorithm integrates an efficient local search procedure associated with the aggressive best-improvement strategy to find high-quality local optimal solutions, a probabilistic criterion to continuously diversify the search and explore new search regions, and a multi-armed bendit algorithm to dynamically select tailored perturbation operators to escape deep local optima. It also takes advantage of an effective TSP heuristic for individual tour optimization.

Computational experiments on two sets of 77 commonly used benchmark instances (with up to 5915 cities and $m=3,5,10,20,30$ routes) show that the proposed algorithm can effectively solve a wide range of instances  in a short running time, leading to new best-known results for 32 challenging instances and matching 35 other best-known results. The algorithm is shown to perform particularly well on instances with 3, 5 and 20 tours, making it an attractive complementary approach to existing algorithms that perform well on instances with 10 tours. These new results can be valuable for future research on the minmax mTSP. In addition, we investigate the contributions of the multi-armed bandit algorithm and the best-improvement strategy for the local search to the performance of the algorithm in the context of the minmax mTSP.

Since the minmax mTSP is a relevant model for a number of real-world problems, our algorithm, whose code will be publicly available, can be used to better solve some of these practical applications.

%In addition, we conduct experment an in-depth examination of the functions and underlying principles governing both the best- and first-improvement strategies employed within the local search procedure. The analysis aims to substantiate the superior effectiveness of the best-improvement strategy in solving routing problems characterized by minmax objectives. Furthermore, we evaluate the effectiveness of the multi-armed bandit algorithm in the context of the minmax mTSP. Since the problem is a relevant model for a number of real-world problems, our algorithm whose code will be publicly available, can be used to better solve some of these practical applications. 

There are several possible directions for future work. First, since the local search component is the most time consuming part of the algorithm, it would be interesting to explore techniques, such as dynamic radius search \cite{gauthier2020inter} to improve its computational efficiency. Second, this work demonstrates the effectiveness of the multi-armed bandit algorithm for the minmax mTSP. This learning technique, combined with various operators, has the potential to be beneficial for other routing problems \cite{maximo2021hybrid,christiaens2020slack,Zheng10259654}. Finally, efficient exact algorithms for the minmax mTSP are still lacking. Research in this area is very valuable. %maximo2021

%\section*{Declaration of competing interest}
%The authors declare that they have no known competing interests that could have appeared to influence the work reported in this work.

\section*{Acknowledgments}
We would like to thank the authors of \cite{ZHENG2022105772}, especially Dr. Jiongzhi Zheng, for sharing the code of their algorithm.

%The authors would like to thank the following colleagues: Dr. F. Arnold and Prof. K. S{\"o}rensen for kindly sharing their source code of the PF algorithm \cite{arnold2021progressive}. %This work is partially supported by the National Natural Science Foundation Program of China (Grant No. 72122006). %Support from the China Scholarship Council (CSC, No. 201906850087) for the first author is also acknowledged.
\ifCLASSOPTIONcaptionsoff
  \newpage
\fi

\bibliographystyle{IEEEtran}
%\bibliographystyle{elsart-num-sort}
%\bibliography{ma_hpmp}
\bibliography{IEEEexample}

\begin{thebibliography}{10}
\providecommand{\url}[1]{#1}
\csname url@samestyle\endcsname
\providecommand{\newblock}{\relax}
\providecommand{\bibinfo}[2]{#2}
\providecommand{\BIBentrySTDinterwordspacing}{\spaceskip=0pt\relax}
\providecommand{\BIBentryALTinterwordstretchfactor}{4}
\providecommand{\BIBentryALTinterwordspacing}{\spaceskip=\fontdimen2\font plus
\BIBentryALTinterwordstretchfactor\fontdimen3\font minus
  \fontdimen4\font\relax}
\providecommand{\BIBforeignlanguage}[2]{{%
\expandafter\ifx\csname l@#1\endcsname\relax
\typeout{** WARNING: IEEEtran.bst: No hyphenation pattern has been}%
\typeout{** loaded for the language `#1'. Using the pattern for}%
\typeout{** the default language instead.}%
\else
\language=\csname l@#1\endcsname
\fi
#2}}
\providecommand{\BIBdecl}{\relax}
\BIBdecl

\bibitem{francca1995m}
P.~M. Fran{\c{c}}a, M.~Gendreau, G.~Laporte, and F.~M. M{\"u}ller, ``The
  m-traveling salesman problem with minmax objective,'' \emph{Transportation
  Science}, vol.~29, no.~3, pp. 267--275, 1995.

\bibitem{dewil2016review}
R.~Dewil, P.~Vansteenwegen, and D.~Cattrysse, ``A review of cutting path
  algorithms for laser cutters,'' \emph{The International Journal of Advanced
  Manufacturing Technology}, vol.~87, pp. 1865--1884, 2016.

\bibitem{zhou2022multi}
B.~Zhou, R.~Zhou, Y.~Gan, F.~Fang, and Y.~Mao, ``Multi-robot multi-station
  cooperative spot welding task allocation based on stepwise optimization: An
  industrial case study,'' \emph{Robotics and Computer-Integrated
  Manufacturing}, vol.~73, p. 102197, 2022.

\bibitem{zhang2018fast}
X.~Zhang and L.~Duan, ``Fast deployment of uav networks for optimal wireless
  coverage,'' \emph{IEEE Transactions on Mobile Computing}, vol.~18, no.~3, pp.
  588--601, 2018.

\bibitem{he2022hybrid}
P.~He and J.-K. Hao, ``Hybrid search with neighborhood reduction for the
  multiple traveling salesman problem,'' \emph{Computers \& Operations
  Research}, vol. 142, p. 105726, 2022.

\bibitem{rao1980note}
M.~Rao, ``A note on the multiple traveling salesmen problem,'' \emph{Operations
  Research}, vol.~28, no. 3-part-i, pp. 628--632, 1980.

\bibitem{karabulut2021modeling}
K.~Karabulut, H.~{\"O}ztop, L.~Kandiller, and M.~F. Tasgetiren, ``Modeling and
  optimization of multiple traveling salesmen problems: An evolution strategy
  approach,'' \emph{Computers \& Operations Research}, vol. 129, p. 105192,
  2021.

\bibitem{ZHENG2022105772}
J.~Zheng, Y.~Hong, W.~Xu, W.~Li, and Y.~Chen, ``An effective iterated two-stage
  heuristic algorithm for the multiple traveling salesmen problem,''
  \emph{Computers \& Operations Research}, vol. 143, p. 105772, 2022.

\bibitem{mahmoudinazlou2023hybrid}
S.~Mahmoudinazlou and C.~Kwon, ``A hybrid genetic algorithm for the min--max
  multiple traveling salesman problem,'' \emph{Computers \& Operations
  Research}, vol. 162, p. 106455, 2024.

\bibitem{he2023Memetic}
P.~He and J.-K. Hao, ``Memetic search for the minmax multiple traveling
  salesman problem with single and multiple depots,'' \emph{European Journal of
  Operational Research}, vol. 307, no.~3, pp. 1055--1070, 2023.

\bibitem{LourencoMS03}
H.~R. Louren{\c{c}}o, O.~C. Martin, and T.~St{\"u}tzle, ``Iterated local
  search,'' in \emph{Handbook of metaheuristics}.\hskip 1em plus 0.5em minus
  0.4em\relax Springer, 2003, pp. 320--353.

\bibitem{ropke2006adaptive}
S.~Ropke and D.~Pisinger, ``An adaptive large neighborhood search heuristic for
  the pickup and delivery problem with time windows,'' \emph{Transportation
  Science}, vol.~40, no.~4, pp. 455--472, 2006.

\bibitem{pisinger2019large}
D.~Pisinger and S.~Ropke, ``Large neighborhood search,'' \emph{Handbook of
  metaheuristics}, pp. 99--127, 2019.

\bibitem{Zhou9729095}
Y.~Zhou, W.~Xu, Z.-H. Fu, and M.~Zhou, ``Multi-neighborhood simulated
  annealing-based iterated local search for colored traveling salesman
  problems,'' \emph{IEEE Transactions on Intelligent Transportation Systems},
  vol.~23, no.~9, pp. 16\,072--16\,082, 2022.

\bibitem{Meng7948721}
X.~Meng, J.~Li, X.~Dai, and J.~Dou, ``Variable neighborhood search for a
  colored traveling salesman problem,'' \emph{IEEE Transactions on Intelligent
  Transportation Systems}, vol.~19, no.~4, pp. 1018--1026, 2018.

\bibitem{nagata2013powerful}
Y.~Nagata and S.~Kobayashi, ``A powerful genetic algorithm using edge assembly
  crossover for the traveling salesman problem,'' \emph{INFORMS Journal on
  Computing}, vol.~25, no.~2, pp. 346--363, 2013.

\bibitem{mladenovic1997variable}
N.~Mladenovi{\'c} and P.~Hansen, ``Variable neighborhood search,''
  \emph{Computers \& Operations Research}, vol.~24, no.~11, pp. 1097--1100,
  1997.

\bibitem{bentley1992fast}
J.~J. Bentley, ``Fast algorithms for geometric traveling salesman problems,''
  \emph{ORSA Journal on Computing}, vol.~4, no.~4, pp. 387--411, 1992.

\bibitem{accorsi2021fast}
L.~Accorsi and D.~Vigo, ``A fast and scalable heuristic for the solution of
  large-scale capacitated vehicle routing problems,'' \emph{Transportation
  Science}, vol.~55, no.~4, pp. 832--856, 2021.

\bibitem{vidal2022hybrid}
T.~Vidal, ``Hybrid genetic search for the {CVRP}: Open-source implementation
  and swap* neighborhood,'' \emph{Computers \& Operations Research}, vol. 140,
  p. 105643, 2022.

\bibitem{he2022general}
P.~He and J.-K. Hao, ``General edge assembly crossover-driven memetic search
  for split delivery vehicle routing,'' \emph{Transportation Science}, vol.~57,
  no.~2, pp. 482--511, 2023.

\bibitem{he2023hybrid}
P.~He, J.-K. Hao, and Q.~Wu, ``A hybrid genetic algorithm for undirected
  traveling salesman problems with profits,'' \emph{Networks: An International
  Journal}, vol.~82, no.~3, pp. 189--221, 2023.

\bibitem{maximo2021hybrid}
V.~R. M{\'a}ximo and M.~C. Nascimento, ``A hybrid adaptive iterated local
  search with diversification control to the capacitated vehicle routing
  problem,'' \emph{European Journal of Operational Research}, vol. 294, no.~3,
  pp. 1108--1119, 2021.

\bibitem{benlic2013breakout}
U.~Benlic and J.-K. Hao, ``Breakout local search for maximum clique problems,''
  \emph{Computers \& Operations Research}, vol.~40, no.~1, pp. 192--206, 2013.

\bibitem{shaw1998using}
P.~Shaw, ``Using constraint programming and local search methods to solve
  vehicle routing problems,'' in \emph{International Conference on Principles
  and Practice of Constraint Programming}.\hskip 1em plus 0.5em minus
  0.4em\relax Springer, 1998, pp. 417--431.

\bibitem{arnold2021pils}
F.~Arnold, {\'I}.~Santana, K.~S{\"o}rensen, and T.~Vidal, ``Pils: exploring
  high-order neighborhoods by pattern mining and injection,'' \emph{Pattern
  Recognition}, vol. 116, p. 107957, 2021.

\bibitem{christiaens2020slack}
J.~Christiaens and G.~Vanden~Berghe, ``Slack induction by string removals for
  vehicle routing problems,'' \emph{Transportation Science}, vol.~54, no.~2,
  pp. 417--433, 2020.

\bibitem{lagos2023multi}
F.~Lagos and J.~Pereira, ``Multi-armed bandit-based hyper-heuristics for
  combinatorial optimization problems,'' \emph{European Journal of Operational
  Research}, vol. 312, no.~1, pp. 70--91, 2024.

\bibitem{venkatesh2015two}
V.~Pandiri and A.~Singh, ``Two metaheuristic approaches for the multiple
  traveling salesperson problem,'' \emph{Applied Soft Computing}, vol.~26, pp.
  74--89, 2015.

\bibitem{wang2017memetic}
Y.~Wang, Y.~Chen, and Y.~Lin, ``Memetic algorithm based on sequential variable
  neighborhood descent for the minmax multiple traveling salesman problem,''
  \emph{Computers \& Industrial Engineering}, vol. 106, pp. 105--122, 2017.

\bibitem{lopez2016irace}
M.~L{\'o}pez-Ib{\'a}{\~n}ez, J.~Dubois-Lacoste, L.~P. C{\'a}ceres,
  M.~Birattari, and T.~St{\"u}tzle, ``The irace package: Iterated racing for
  automatic algorithm configuration,'' \emph{Operations Research Perspectives},
  vol.~3, pp. 43--58, 2016.

\bibitem{gauthier2020inter}
J.~B. Gauthier and S.~Irnich, ``Inter-depot moves and dynamic-radius search for
  multi-depot vehicle routing problems,'' \emph{Gutenberg School of Management
  and Economics (2022)}, 2020.

\bibitem{Zheng10259654}
Z.~Zheng, S.~Yao, G.~Li, L.~Han, and Z.~Wang, ``Pareto improver: Learning
  improvement heuristics for multi-objective route planning,'' \emph{IEEE
  Transactions on Intelligent Transportation Systems}, vol.~25, no.~1, pp.
  1033--1043, 2024.

\bibitem{bektas2006multiple}
T.~Bektas, ``The multiple traveling salesman problem: an overview of
  formulations and solution procedures,'' \emph{Omega}, vol.~34, no.~3, pp.
  209--219, 2006.

\end{thebibliography}

\appendix{}

\textbf{A. Mathematical model}\label{math_model}\\
%\noindent \textbf{Mathematical model}

In this section, a flow-based formulation for the minmax mTSP is provided by \cite{bektas2006multiple}. For a directed graph $\mathcal{G}=(\mathcal{V},\mathcal{A})$, \textit{in-arcs} and \textit{out-arcs} of the vertex set $\mathcal{W}$ are defined as $\delta^{-}(\mathcal{W})=\{(i,j)\in\mathcal{A}:i\notin\mathcal{W},j\in\mathcal{W}\}$ and $\delta^{+}(\mathcal{W})=\{(i,j)\in\mathcal{A}:i\in\mathcal{W},j\notin\mathcal{W}\}$, respectively. It has become a standard to define $\delta(i):=\delta(\{i\})$ for singleton set $\mathcal{W}=\{i\}$ (similarly, $\delta^{+}(i)$ and $\delta^{-}(i)$). Let $C$ be an auxiliary variable to measure the maximum cost among $m$ tours. Let $x_{ijk}$ be a binary variable that equals 1 if arc $(i,j)$ is used in the solution, 0 otherwise. Let $u_i$ be an auxiliary that denotes the rank of vertex $i\in\mathcal{V}$ in the order of visits. Then $(\sum_{k\in \mathcal{M}}{x_{ijk}} = 1)\to (u_i+1 = u_j)$. The flow-based formulation for the minmax mTSP is as follows:

\begin{equation}\label{math_obj}
Min\   C
\end{equation}
\ \ \ \ \ \ \ \ \ \ \ \ \ \ subject to:
\begin{equation}\label{math_eq1}
\sum_{i\in \mathcal{V}}{\sum_{j\in \mathcal{V}}c_{ij}x_{ijk}} \leq C \ \ \ (\forall k\in \mathcal{M})
\end{equation}
\begin{equation}\label{math_eq2}
\sum_{k\in\mathcal{M}}{\sum_{i\in\delta^{+}(j)}{x_{ijk} = 1}} \ \ \  (\forall j\in \mathcal{N})
\end{equation}
\begin{equation}\label{math_eq3}
\sum_{j\in\delta^{+}(v_0)}{x_{v_0jk} = 1} \ \ \  (\forall k\in \mathcal{M})
\end{equation}
\begin{equation}\label{math_eq4}
\sum_{i\in\delta^{-}(j)}{x_{ijk}} - \sum_{i\in\delta^{+}(j)}{x_{ijk}} = 0 \ \ \  (\forall j\in \mathcal{N}, k\in \mathcal{M})
\end{equation}
\begin{equation}\label{math_eq5}
\sum_{i\in\delta^{-}(v_0)}{x_{iv_0k} = 1} \ \ \  (\forall k\in \mathcal{M})
\end{equation}
\begin{equation}\label{math_eq6}
u_i - u_j + |\mathcal{V}|\sum_{k\in\mathcal{M}}x_{ijk} \leq |\mathcal{V}|-1 \ \ \  (\forall i\neq j, i,j\in\mathcal{N})
\end{equation}
\begin{equation}\label{math_eq7}
x_{ijk}\in\{0,1\} \ \ \  (\forall k\in \mathcal{M}, (i,j)\in\mathcal{A})
\end{equation}

Objective function \ref{math_obj} and constraints \ref{math_eq1} aim at minimizing the longest tour among $m$ tours. Constraints \ref{math_eq2} ensure that each city is assigned to exactly one tour. Next, constraints \ref{math_eq3}-\ref{math_eq5} define a source-to-sink path in graph $\mathcal{G}$ for each tour $k$. Constraints \ref{math_eq6} are the extensions of MTZ-based formulation for the TSP.\\

\noindent\textbf{B. Computational results}\label{comput_results}\\

%\section{Computational results}\label{comput_results}

This section presents the detailed computational results of the proposed MILS algorithm together with the results of reference algorithms. In Tables \ref{tableA1} and \ref{tableA2}, column 'Instance' indicates the name of each instance (Instances marked with an asterisk * have proven optimal solutions); column 'BKS' is the best-known results summarized from the literature; 'Best' and 'Avg. ' are the best and average results over 20 independent runs, respectively, obtained by the corresponding algorithm in the column header; '$\delta$(\%)' is calculated as $\delta=100 \times (f_{best}-BKS)/BKS$, where $f_{best}$ is the best objective value of MILS. The \textit{Average} row is the average value of a performance indicator over the instances of a benchmark set. Improved best results (new upper bounds) are indicated by negative $\delta$(\%) values highlighted in bold. In all tables, the dark gray color indicates that the corresponding algorithm obtains the best result among the compared algorithms on the corresponding instance; the medium gray color displays the second best results, and so on.

\newpage

\begin{table*}[h]
\caption{Results for the minmax mTSP on the instances of set $\mathbb{S}$.} \label{tableA1}
\renewcommand{\baselinestretch}{1.1}
%\vskip 0.15in
\begin{center}
\begin{tiny}
\begin{tabular}{p{0.9cm}p{0.5cm}p{0.6cm}p{0.8cm}p{0.7cm}p{0.7cm}p{0.6cm}p{0.6cm}p{0.01cm}p{0.8cm}p{0.8cm}p{0.7cm}p{0.7cm}p{0.7cm}}
\hline
\multicolumn{1}{c}{}                            & \multicolumn{1}{c}{}                      & \multicolumn{6}{c}{Best}                                                                                                                                                                      &  & \multicolumn{5}{c}{Avg.}                                                                                                                                                     \\
\cline{3-8} \cline{10-14}
\multicolumn{1}{c}{\multirow{-2}{*}{Instances}} & \multicolumn{1}{c}{\multirow{-2}{*}{BKS}} & HSNR\cite{he2022hybrid}                             & ITSHA \cite{ZHENG2022105772}                            & HGA \cite{mahmoudinazlou2023hybrid}                              & MA \cite{he2023Memetic}                               & MILS                             & $\delta$(\%)        &  & HSNR \cite{he2022hybrid}                             & ITSHA \cite{ZHENG2022105772}                           & HGA \cite{mahmoudinazlou2023hybrid}                              & MA \cite{he2023Memetic}                               & MILS                             \\
\hline
mtsp51-3   & 159.57   & \cellcolor[HTML]{A6A6A6}159.57   & \cellcolor[HTML]{A6A6A6}159.57   & \cellcolor[HTML]{A6A6A6}159.57   & \cellcolor[HTML]{A6A6A6}159.57   & \cellcolor[HTML]{A6A6A6}159.57   & 0.00           &  & 159.85                           & \cellcolor[HTML]{A6A6A6}159.57   & \cellcolor[HTML]{A6A6A6}159.57   & \cellcolor[HTML]{A6A6A6}159.57   & \cellcolor[HTML]{A6A6A6}159.57   \\
mtsp51-5   & 118.13   & \cellcolor[HTML]{A6A6A6}118.13   & \cellcolor[HTML]{A6A6A6}118.13   & \cellcolor[HTML]{A6A6A6}118.13   & \cellcolor[HTML]{A6A6A6}118.13   & \cellcolor[HTML]{A6A6A6}118.13   & 0.00           &  & \cellcolor[HTML]{A6A6A6}118.13   & \cellcolor[HTML]{A6A6A6}118.13   & \cellcolor[HTML]{A6A6A6}118.13   & \cellcolor[HTML]{A6A6A6}118.13   & \cellcolor[HTML]{A6A6A6}118.13   \\
mtsp51-10  & 112.07*  & \cellcolor[HTML]{A6A6A6}112.07   & \cellcolor[HTML]{A6A6A6}112.07   & \cellcolor[HTML]{A6A6A6}112.07   & \cellcolor[HTML]{A6A6A6}112.07   & \cellcolor[HTML]{A6A6A6}112.07   & 0.00           &  & \cellcolor[HTML]{A6A6A6}112.07   & \cellcolor[HTML]{A6A6A6}112.07   & \cellcolor[HTML]{A6A6A6}112.07   & \cellcolor[HTML]{A6A6A6}112.07   & \cellcolor[HTML]{A6A6A6}112.07   \\
mtsp100-3  & 8509.16  & \cellcolor[HTML]{A6A6A6}8509.16  & \cellcolor[HTML]{A6A6A6}8509.16  & \cellcolor[HTML]{A6A6A6}8509.16  & \cellcolor[HTML]{A6A6A6}8509.16  & \cellcolor[HTML]{A6A6A6}8509.16  & 0.00           &  & 8513.75                          & \cellcolor[HTML]{A6A6A6}8509.16  & \cellcolor[HTML]{A6A6A6}8509.16  & \cellcolor[HTML]{A6A6A6}8509.16  & \cellcolor[HTML]{A6A6A6}8509.16  \\
mtsp100-5  & 6765.73  & \cellcolor[HTML]{A6A6A6}6765.73  & \cellcolor[HTML]{FFFFFF}6767.82  & \cellcolor[HTML]{FFFFFF}6767.82  & \cellcolor[HTML]{A6A6A6}6765.73  & \cellcolor[HTML]{D9D9D9}6766.73  & 0.01           &  & \cellcolor[HTML]{F2F2F2}6770.67  & 6772.95                          & \cellcolor[HTML]{D9D9D9}6770.50  & \cellcolor[HTML]{A6A6A6}6766.78  & \cellcolor[HTML]{BFBFBF}6768.16  \\
mtsp100-10 & 6358.49* & \cellcolor[HTML]{A6A6A6}6358.49  & \cellcolor[HTML]{A6A6A6}6358.49  & \cellcolor[HTML]{A6A6A6}6358.49  & \cellcolor[HTML]{A6A6A6}6358.49  & \cellcolor[HTML]{A6A6A6}6358.49  & 0.00           &  & \cellcolor[HTML]{A6A6A6}6358.49  & \cellcolor[HTML]{A6A6A6}6358.49  & \cellcolor[HTML]{A6A6A6}6358.49  & \cellcolor[HTML]{A6A6A6}6358.49  & \cellcolor[HTML]{A6A6A6}6358.49  \\
mtsp100-20 & 6358.49* & \cellcolor[HTML]{A6A6A6}6358.49  & \cellcolor[HTML]{A6A6A6}6358.49  & \cellcolor[HTML]{A6A6A6}6358.49  & \cellcolor[HTML]{A6A6A6}6358.49  & \cellcolor[HTML]{A6A6A6}6358.49  & 0.00           &  & \cellcolor[HTML]{A6A6A6}6358.49  & \cellcolor[HTML]{A6A6A6}6358.49  & \cellcolor[HTML]{A6A6A6}6358.49  & \cellcolor[HTML]{A6A6A6}6358.49  & \cellcolor[HTML]{A6A6A6}6358.49  \\
rand100-3  & 3031.95  & \cellcolor[HTML]{A6A6A6}3031.95  & \cellcolor[HTML]{A6A6A6}3031.95  & \cellcolor[HTML]{A6A6A6}3031.95  & \cellcolor[HTML]{A6A6A6}3031.95  & \cellcolor[HTML]{A6A6A6}3031.95  & 0.00           &  & 3032.67                          & 3033.65                          & \cellcolor[HTML]{A6A6A6}3031.95  & \cellcolor[HTML]{A6A6A6}3031.95  & \cellcolor[HTML]{A6A6A6}3031.95  \\
rand100-5  & 2409.63  & 2411.68                          & \cellcolor[HTML]{D9D9D9}2412.35  & \cellcolor[HTML]{D9D9D9}2412.35  & \cellcolor[HTML]{A6A6A6}2409.63  & \cellcolor[HTML]{A6A6A6}2409.63  & 0.00           &  & 2415.00                          & \cellcolor[HTML]{F2F2F2}2414.65  & \cellcolor[HTML]{A6A6A6}2409.63  & \cellcolor[HTML]{BFBFBF}2409.64  & \cellcolor[HTML]{D9D9D9}2409.96  \\
rand100-10 & 2299.16* & \cellcolor[HTML]{A6A6A6}2299.16  & \cellcolor[HTML]{A6A6A6}2299.16  & \cellcolor[HTML]{A6A6A6}2299.16  & \cellcolor[HTML]{A6A6A6}2299.16  & \cellcolor[HTML]{A6A6A6}2299.16  & 0.00           &  & \cellcolor[HTML]{A6A6A6}2299.16  & \cellcolor[HTML]{A6A6A6}2299.16  & \cellcolor[HTML]{A6A6A6}2299.16  & \cellcolor[HTML]{A6A6A6}2299.16  & \cellcolor[HTML]{A6A6A6}2299.16  \\
rand100-20 & 2299.16* & \cellcolor[HTML]{A6A6A6}2299.16  & \cellcolor[HTML]{A6A6A6}2299.16  & \cellcolor[HTML]{A6A6A6}2299.16  & \cellcolor[HTML]{A6A6A6}2299.16  & \cellcolor[HTML]{A6A6A6}2299.16  & 0.00           &  & \cellcolor[HTML]{A6A6A6}2299.16  & \cellcolor[HTML]{A6A6A6}2299.16  & \cellcolor[HTML]{A6A6A6}2299.16  & \cellcolor[HTML]{A6A6A6}2299.16  & \cellcolor[HTML]{A6A6A6}2299.16  \\
mtsp150-3  & 13038.30 & \cellcolor[HTML]{F2F2F2}13075.80 & 13088.74                         & \cellcolor[HTML]{D9D9D9}13038.34 & \cellcolor[HTML]{A6A6A6}13038.30 & \cellcolor[HTML]{A6A6A6}13038.30 & 0.00           &  & \cellcolor[HTML]{F2F2F2}13169.37 & 13210.69                         & \cellcolor[HTML]{D9D9D9}13093.48 & \cellcolor[HTML]{BFBFBF}13079.74 & \cellcolor[HTML]{A6A6A6}13038.46 \\
mtsp150-5  & 8417.02  & 8477.96                          & 8492.97                          & \cellcolor[HTML]{A6A6A6}8417.02  & \cellcolor[HTML]{A6A6A6}8417.02  & \cellcolor[HTML]{A6A6A6}8417.02  & 0.00           &  & \cellcolor[HTML]{F2F2F2}8538.83  & 8572.77                          & \cellcolor[HTML]{D9D9D9}8486.81  & \cellcolor[HTML]{BFBFBF}8453.15  & \cellcolor[HTML]{A6A6A6}8430.47  \\
mtsp150-10 & 5557.41  & \cellcolor[HTML]{F2F2F2}5590.64  & 5593.56                          & \cellcolor[HTML]{BFBFBF}5561.97  & \cellcolor[HTML]{A6A6A6}5557.41  & \cellcolor[HTML]{D9D9D9}5590.19  & 0.59           &  & \cellcolor[HTML]{D9D9D9}5604.92  & 5608.50                          & \cellcolor[HTML]{A6A6A6}5587.37  & \cellcolor[HTML]{BFBFBF}5588.86  & \cellcolor[HTML]{F2F2F2}5607.72  \\
mtsp150-20 & 5246.49* & \cellcolor[HTML]{A6A6A6}5246.49  & \cellcolor[HTML]{A6A6A6}5246.49  & \cellcolor[HTML]{A6A6A6}5246.49  & \cellcolor[HTML]{A6A6A6}5246.49  & \cellcolor[HTML]{A6A6A6}5246.49  & 0.00           &  & \cellcolor[HTML]{A6A6A6}5246.49  & \cellcolor[HTML]{A6A6A6}5246.49  & \cellcolor[HTML]{A6A6A6}5246.49  & \cellcolor[HTML]{A6A6A6}5246.49  & \cellcolor[HTML]{A6A6A6}5246.49  \\
mtsp150-30 & 5246.49* & \cellcolor[HTML]{A6A6A6}5246.49  & \cellcolor[HTML]{A6A6A6}5246.49  & -                                & \cellcolor[HTML]{A6A6A6}5246.49  & \cellcolor[HTML]{A6A6A6}5246.49  & 0.00           &  & \cellcolor[HTML]{A6A6A6}5246.49  & \cellcolor[HTML]{A6A6A6}5246.49  & \cellcolor[HTML]{A6A6A6}-        & \cellcolor[HTML]{A6A6A6}5246.49  & \cellcolor[HTML]{A6A6A6}5246.49  \\
gtsp150-3  & 2401.63  & \cellcolor[HTML]{F2F2F2}2407.34  & \cellcolor[HTML]{F2F2F2}2407.34  & \cellcolor[HTML]{A6A6A6}2401.63  & \cellcolor[HTML]{A6A6A6}2401.63  & \cellcolor[HTML]{A6A6A6}2401.63  & 0.00           &  & 2435.49                          & \cellcolor[HTML]{F2F2F2}2416.87  & \cellcolor[HTML]{A6A6A6}2401.63  & \cellcolor[HTML]{BFBFBF}2401.86  & \cellcolor[HTML]{D9D9D9}2402.19  \\
gtsp150-5  & 1741.13  & 1741.71                          & \cellcolor[HTML]{A6A6A6}1741.13  & \cellcolor[HTML]{A6A6A6}1741.13  & \cellcolor[HTML]{A6A6A6}1741.13  & \cellcolor[HTML]{A6A6A6}1741.13  & 0.00           &  & 1743.48                          & 1752.06                          & \cellcolor[HTML]{A6A6A6}1741.13  & \cellcolor[HTML]{A6A6A6}1741.13  & \cellcolor[HTML]{A6A6A6}1741.13  \\
gtsp150-10 & 1554.64* & \cellcolor[HTML]{A6A6A6}1554.64  & \cellcolor[HTML]{A6A6A6}1554.64  & \cellcolor[HTML]{A6A6A6}1554.64  & \cellcolor[HTML]{A6A6A6}1554.64  & \cellcolor[HTML]{A6A6A6}1554.64  & 0.00           &  & \cellcolor[HTML]{A6A6A6}1554.64  & \cellcolor[HTML]{A6A6A6}1554.64  & \cellcolor[HTML]{A6A6A6}1554.64  & \cellcolor[HTML]{F2F2F2}1554.76  & 1557.42                          \\
gtsp150-20 & 1554.64* & \cellcolor[HTML]{A6A6A6}1554.64  & \cellcolor[HTML]{A6A6A6}1554.64  & \cellcolor[HTML]{A6A6A6}1554.64  & \cellcolor[HTML]{A6A6A6}1554.64  & \cellcolor[HTML]{A6A6A6}1554.64  & 0.00           &  & \cellcolor[HTML]{A6A6A6}1554.64  & \cellcolor[HTML]{A6A6A6}1554.64  & \cellcolor[HTML]{A6A6A6}1554.64  & \cellcolor[HTML]{A6A6A6}1554.64  & \cellcolor[HTML]{A6A6A6}1554.64  \\
gtsp150-30 & 1554.64* & \cellcolor[HTML]{A6A6A6}1554.64  & \cellcolor[HTML]{A6A6A6}1554.64  & -                                & \cellcolor[HTML]{A6A6A6}1554.64  & \cellcolor[HTML]{A6A6A6}1554.64  & 0.00           &  & \cellcolor[HTML]{A6A6A6}1554.64  & \cellcolor[HTML]{A6A6A6}1554.64  & \cellcolor[HTML]{A6A6A6}-        & \cellcolor[HTML]{A6A6A6}1554.64  & \cellcolor[HTML]{A6A6A6}1554.64  \\
kroA200-3  & 10691.00 & 10748.10                         & \cellcolor[HTML]{F2F2F2}10700.57 & \cellcolor[HTML]{D9D9D9}10691.03 & \cellcolor[HTML]{A6A6A6}10691.00 & \cellcolor[HTML]{A6A6A6}10691.00 & 0.00           &  & 10987.69                         & \cellcolor[HTML]{F2F2F2}10819.85 & \cellcolor[HTML]{D9D9D9}10700.06 & \cellcolor[HTML]{BFBFBF}10691.41 & \cellcolor[HTML]{A6A6A6}10691.00 \\
kroA200-5  & 7412.12  & \cellcolor[HTML]{F2F2F2}7418.87  & 7449.22                          & \cellcolor[HTML]{F2F2F2}7421.12  & \cellcolor[HTML]{A6A6A6}7412.12  & \cellcolor[HTML]{BFBFBF}7413.80  & 0.02           &  & \cellcolor[HTML]{F2F2F2}7494.44  & 7513.67                          & \cellcolor[HTML]{D9D9D9}7420.23  & \cellcolor[HTML]{A6A6A6}7414.21  & \cellcolor[HTML]{BFBFBF}7418.57  \\
kroA200-10 & 6223.22* & \cellcolor[HTML]{A6A6A6}6223.22  & \cellcolor[HTML]{A6A6A6}6223.22  & \cellcolor[HTML]{A6A6A6}6223.22  & \cellcolor[HTML]{A6A6A6}6223.22  & \cellcolor[HTML]{A6A6A6}6223.22  & 0.00           &  & \cellcolor[HTML]{A6A6A6}6223.22  & \cellcolor[HTML]{A6A6A6}6223.22  & \cellcolor[HTML]{A6A6A6}6223.22  & 6249.10                          & \cellcolor[HTML]{F2F2F2}6226.12  \\
kroA200-20 & 6223.22* & \cellcolor[HTML]{A6A6A6}6223.22  & \cellcolor[HTML]{A6A6A6}6223.22  & \cellcolor[HTML]{A6A6A6}6223.22  & \cellcolor[HTML]{A6A6A6}6223.22  & \cellcolor[HTML]{A6A6A6}6223.22  & 0.00           &  & \cellcolor[HTML]{A6A6A6}6223.22  & \cellcolor[HTML]{A6A6A6}6223.22  & \cellcolor[HTML]{A6A6A6}6223.22  & \cellcolor[HTML]{A6A6A6}6223.22  & \cellcolor[HTML]{A6A6A6}6223.22  \\
lin318-3   & 15663.50 & \cellcolor[HTML]{F2F2F2}15902.50 & 15930.04                         & \cellcolor[HTML]{D9D9D9}15698.61 & \cellcolor[HTML]{A6A6A6}15663.50 & \cellcolor[HTML]{A6A6A6}15663.50 & 0.00           &  & 16207.05                         & \cellcolor[HTML]{F2F2F2}16088.56 & \cellcolor[HTML]{BFBFBF}15714.31 & \cellcolor[HTML]{A6A6A6}15699.92 & \cellcolor[HTML]{D9D9D9}15751.70 \\
lin318-5   & 11276.80 & \cellcolor[HTML]{F2F2F2}11295.20 & 11430.65                         & \cellcolor[HTML]{D9D9D9}11289.26 & \cellcolor[HTML]{A6A6A6}11276.80 & \cellcolor[HTML]{A6A6A6}11276.80 & 0.00           &  & \cellcolor[HTML]{F2F2F2}11596.35 & 11601.67                         & \cellcolor[HTML]{BFBFBF}11297.35 & \cellcolor[HTML]{A6A6A6}11291.59 & \cellcolor[HTML]{D9D9D9}11301.19 \\
lin318-10  & 9731.17* & \cellcolor[HTML]{A6A6A6}9731.17  & \cellcolor[HTML]{A6A6A6}9731.17  & \cellcolor[HTML]{A6A6A6}9731.17  & \cellcolor[HTML]{A6A6A6}9731.17  & \cellcolor[HTML]{A6A6A6}9731.17  & 0.00           &  & \cellcolor[HTML]{A6A6A6}9731.17  & \cellcolor[HTML]{A6A6A6}9731.17  & \cellcolor[HTML]{A6A6A6}9731.17  & \cellcolor[HTML]{A6A6A6}9731.17  & 9731.29                          \\
lin318-20  & 9731.17* & \cellcolor[HTML]{A6A6A6}9731.17  & \cellcolor[HTML]{A6A6A6}9731.17  & \cellcolor[HTML]{A6A6A6}9731.17  & \cellcolor[HTML]{A6A6A6}9731.17  & \cellcolor[HTML]{A6A6A6}9731.17  & 0.00           &  & \cellcolor[HTML]{A6A6A6}9731.17  & \cellcolor[HTML]{A6A6A6}9731.17  & \cellcolor[HTML]{A6A6A6}9731.17  & \cellcolor[HTML]{A6A6A6}9731.17  & \cellcolor[HTML]{A6A6A6}9731.17  \\
att532-3   & 9966.00  & 10231.00                         & \cellcolor[HTML]{F2F2F2}10158.00 & \cellcolor[HTML]{D9D9D9}9973.00  & \cellcolor[HTML]{BFBFBF}9966.00  & \cellcolor[HTML]{A6A6A6}9926.00  & \textbf{-0.40} &  & 10565.30                         & \cellcolor[HTML]{F2F2F2}10344.50 & \cellcolor[HTML]{BFBFBF}10038.03 & \cellcolor[HTML]{D9D9D9}10064.00 & \cellcolor[HTML]{A6A6A6}9938.25  \\
att532-5   & 6950.00  & 7067.00                          & 7067.00                          & \cellcolor[HTML]{BFBFBF}6950.00  & \cellcolor[HTML]{D9D9D9}6986.00  & \cellcolor[HTML]{A6A6A6}6942.00  & \textbf{-0.12} &  & 7334.00                          & \cellcolor[HTML]{F2F2F2}7156.80  & \cellcolor[HTML]{BFBFBF}7003.80  & \cellcolor[HTML]{D9D9D9}7070.95  & \cellcolor[HTML]{A6A6A6}6976.35  \\
att532-10  & 5698.00  & \cellcolor[HTML]{BFBFBF}5709.00  & \cellcolor[HTML]{D9D9D9}5731.00  & \cellcolor[HTML]{A6A6A6}5698.00  & \cellcolor[HTML]{F2F2F2}5770.00  & 5783.00                          & 1.49           &  & \cellcolor[HTML]{BFBFBF}5738.90  & \cellcolor[HTML]{D9D9D9}5787.50  & \cellcolor[HTML]{A6A6A6}5716.75  & \cellcolor[HTML]{F2F2F2}5796.75  & 5846.30                          \\
att532-20  & 5580*    & \cellcolor[HTML]{A6A6A6}5580.00  & \cellcolor[HTML]{A6A6A6}5583.00  & \cellcolor[HTML]{A6A6A6}5580.00  & \cellcolor[HTML]{A6A6A6}5580.00  & \cellcolor[HTML]{A6A6A6}5580.00  & 0.00           &  & \cellcolor[HTML]{A6A6A6}5580.00  & \cellcolor[HTML]{F2F2F2}5601.75  & \cellcolor[HTML]{A6A6A6}5580.00  & \cellcolor[HTML]{D9D9D9}5589.35  & \cellcolor[HTML]{A6A6A6}5580.00  \\
rat783-3   & 3052.41  & 3187.90                          & \cellcolor[HTML]{F2F2F2}3131.99  & \cellcolor[HTML]{D9D9D9}3086.70  & \cellcolor[HTML]{BFBFBF}3052.41  & \cellcolor[HTML]{A6A6A6}3040.75  & \textbf{-0.38} &  & 3237.29                          & \cellcolor[HTML]{F2F2F2}3180.79  & \cellcolor[HTML]{D9D9D9}3106.39  & \cellcolor[HTML]{BFBFBF}3083.52  & \cellcolor[HTML]{A6A6A6}3052.02  \\
rat783-5   & 1959.16  & \cellcolor[HTML]{F2F2F2}2006.46  & 2018.44                          & \cellcolor[HTML]{BFBFBF}1959.16  & \cellcolor[HTML]{D9D9D9}1961.12  & \cellcolor[HTML]{A6A6A6}1941.90  & \textbf{-0.88} &  & \cellcolor[HTML]{F2F2F2}2044.32  & 2058.65                          & \cellcolor[HTML]{BFBFBF}1981.51  & \cellcolor[HTML]{D9D9D9}1989.68  & \cellcolor[HTML]{A6A6A6}1948.48  \\
rat783-10  & 1304.70  & \cellcolor[HTML]{F2F2F2}1334.76  & 1357.65                          & \cellcolor[HTML]{A6A6A6}1304.70  & \cellcolor[HTML]{BFBFBF}1313.01  & \cellcolor[HTML]{D9D9D9}1318.77  & 1.08           &  & \cellcolor[HTML]{F2F2F2}1345.88  & 1381.69                          & \cellcolor[HTML]{A6A6A6}1321.33  & \cellcolor[HTML]{BFBFBF}1325.54  & \cellcolor[HTML]{D9D9D9}1341.19  \\
rat783-20  & 1231.69* & \cellcolor[HTML]{A6A6A6}1231.69  & \cellcolor[HTML]{A6A6A6}1231.69  & \cellcolor[HTML]{A6A6A6}1231.69  & \cellcolor[HTML]{A6A6A6}1231.69  & \cellcolor[HTML]{A6A6A6}1231.69  & 0.00           &  & \cellcolor[HTML]{A6A6A6}1231.69  & \cellcolor[HTML]{D9D9D9}1231.84  & \cellcolor[HTML]{A6A6A6}1231.69  & \cellcolor[HTML]{F2F2F2}1235.37  & \cellcolor[HTML]{A6A6A6}1231.69  \\
pcb1173-3  & 19569.50 & 20813.80                         & \cellcolor[HTML]{F2F2F2}20288.75 & \cellcolor[HTML]{D9D9D9}19724.24 & \cellcolor[HTML]{BFBFBF}19569.50 & \cellcolor[HTML]{A6A6A6}19412.40 & \textbf{-0.80} &  & 21144.92                         & \cellcolor[HTML]{F2F2F2}20473.45 & \cellcolor[HTML]{D9D9D9}20016.94 & \cellcolor[HTML]{BFBFBF}19858.77 & \cellcolor[HTML]{A6A6A6}19572.01 \\
pcb1173-5  & 12406.60 & 13032.30                         & 12816.55                         & 12517.65                         & \cellcolor[HTML]{BFBFBF}12406.60 & \cellcolor[HTML]{A6A6A6}12224.60 & \textbf{-1.47} &  & 13216.99                         & \cellcolor[HTML]{F2F2F2}13045.14 & \cellcolor[HTML]{D9D9D9}12655.64 & \cellcolor[HTML]{BFBFBF}12639.49 & \cellcolor[HTML]{A6A6A6}12266.15 \\
pcb1173-10 & 7512.43  & \cellcolor[HTML]{F2F2F2}7758.26  & 7801.18                          & \cellcolor[HTML]{BFBFBF}7512.43  & \cellcolor[HTML]{D9D9D9}7623.59  & \cellcolor[HTML]{A6A6A6}7476.78  & \textbf{-0.47} &  & \cellcolor[HTML]{F2F2F2}7897.20  & 7910.09                          & \cellcolor[HTML]{BFBFBF}7617.21  & \cellcolor[HTML]{D9D9D9}7745.00  & \cellcolor[HTML]{A6A6A6}7511.91  \\
pcb1173-20 & 6528.86* & \cellcolor[HTML]{A6A6A6}6528.86  & \cellcolor[HTML]{A6A6A6}6528.86  & \cellcolor[HTML]{A6A6A6}6528.86  & \cellcolor[HTML]{A6A6A6}6528.86  & \cellcolor[HTML]{A6A6A6}6528.86  & 0.00           &  & \cellcolor[HTML]{A6A6A6}6528.86  & \cellcolor[HTML]{D9D9D9}6534.75  & \cellcolor[HTML]{A6A6A6}6528.86  & \cellcolor[HTML]{F2F2F2}6548.87  & \cellcolor[HTML]{A6A6A6}6528.86  \\
\hline
Average    & 6900.50  & 6015.33                          & 6000.98                          & 6077.37                          & 5943.29                          & 5929.96                          & -              &  & 6076.73                          & 6043.73                          & 6100.80                          & 5971.30                          & 5943.21                         
            \\                     
\hline               
\end{tabular}
\end{tiny}
\end{center}
%\vskip -0.10in
\end{table*}
%\end{table}
\renewcommand{\baselinestretch}{1.0}\large\normalsize

%\begin{sidewaystable}[htp]
\begin{table*}[h]
\caption{Results for the minmax mTSP on the instances of set $\mathbb{L}$.} \label{tableA2}
\renewcommand{\baselinestretch}{1.1}
%\vskip 0.15in
\begin{center}
\begin{tiny}
\begin{tabular}{p{1.0cm}p{0.8cm}p{0.8cm}p{0.8cm}p{0.8cm}p{0.8cm}p{0.6cm}p{0.01cm}p{0.8cm}p{0.8cm}p{0.8cm}p{0.8cm}}
\hline
\multicolumn{1}{c}{}                            & \multicolumn{1}{c}{}                      & \multicolumn{5}{c}{Best}                                                                                                                                      &                         & \multicolumn{4}{c}{Avg.}                                                                                                                     \\
\cline{3-7}\cline{9-12}
\multicolumn{1}{c}{\multirow{-2}{*}{Instances}} & \multicolumn{1}{c}{\multirow{-2}{*}{BKS}} & HSNR \cite{he2022hybrid}                            & ITSHA \cite{ZHENG2022105772}                             & MA \cite{he2023Memetic}                                & MILS                              & $\delta$(\%)        &                         & HSNR \cite{he2022hybrid}                            & ITSHA \cite{ZHENG2022105772}                            & MA \cite{he2023Memetic}                               & MILS                              \\
\hline
nrw1379-3                                       & 19222.10                                  & 20495.90                         & \cellcolor[HTML]{F2F2F2}19871.21  & \cellcolor[HTML]{D9D9D9}19222.10  & \cellcolor[HTML]{BFBFBF}19126.40  & \textbf{-0.50} &                         & 20765.70                         & \cellcolor[HTML]{F2F2F2}20085.29  & \cellcolor[HTML]{D9D9D9}19472.39  & \cellcolor[HTML]{BFBFBF}19202.46  \\
nrw1379-5                                       & 11913.40                                  & 12416.50                         & \cellcolor[HTML]{F2F2F2}12218.09  & \cellcolor[HTML]{D9D9D9}11913.40  & \cellcolor[HTML]{BFBFBF}11729.70  & \textbf{-1.54} &                         & 12652.56                         & \cellcolor[HTML]{F2F2F2}12493.99  & \cellcolor[HTML]{D9D9D9}12203.76  & \cellcolor[HTML]{BFBFBF}11769.61  \\
nrw1379-10                                      & 6846.27                                   & 7114.71                          & 7008.61                           & 6846.27                           & \cellcolor[HTML]{BFBFBF}6576.46   & \textbf{-3.94} &                         & 7212.24                          & \cellcolor[HTML]{F2F2F2}7136.56   & \cellcolor[HTML]{D9D9D9}6987.31   & \cellcolor[HTML]{BFBFBF}6626.48   \\
nrw1379-20                                      & 5370.82*                                  & \cellcolor[HTML]{BFBFBF}5370.82  & \cellcolor[HTML]{BFBFBF}5370.82   & \cellcolor[HTML]{F2F2F2}5381.59   & \cellcolor[HTML]{BFBFBF}5370.82   & 0.00           &  & \cellcolor[HTML]{BFBFBF}5371.08  & \cellcolor[HTML]{F2F2F2}5402.42   & \cellcolor[HTML]{D9D9D9}5388.99   & \cellcolor[HTML]{BFBFBF}5371.08   \\

fl1400-3                                        & 7854.97                                   & 9192.38                          & 8310.34                           & \cellcolor[HTML]{BFBFBF}7854.97   & \cellcolor[HTML]{D9D9D9}7894.73   & 0.51           &                         & 9621.59                          & \cellcolor[HTML]{F2F2F2}8482.77   & \cellcolor[HTML]{BFBFBF}7989.25   & \cellcolor[HTML]{D9D9D9}8137.51   \\
fl1400-5                                        & 6116.28                                   & \cellcolor[HTML]{D9D9D9}6268.25  & \cellcolor[HTML]{F2F2F2}6360.45   & \cellcolor[HTML]{BFBFBF}6116.28   & \cellcolor[HTML]{FFFFFF}6280.38   & 2.68           &                         & 6783.62                          & \cellcolor[HTML]{D9D9D9}6572.53   & \cellcolor[HTML]{BFBFBF}6185.40   & \cellcolor[HTML]{F2F2F2}6701.72   \\
fl1400-10                                       & 5763.26*                                  & \cellcolor[HTML]{BFBFBF}5763.26  & \cellcolor[HTML]{BFBFBF}5763.26   & \cellcolor[HTML]{BFBFBF}5763.26   & \cellcolor[HTML]{BFBFBF}5763.26   & 0.00           &                         & \cellcolor[HTML]{BFBFBF}5763.26  & 5785.69                           & \cellcolor[HTML]{BFBFBF}5763.26   & \cellcolor[HTML]{BFBFBF}5763.26   \\
fl1400-20                                       & 5763.26*                                  & \cellcolor[HTML]{BFBFBF}5763.26  & \cellcolor[HTML]{BFBFBF}5763.26   & \cellcolor[HTML]{BFBFBF}5763.26   & \cellcolor[HTML]{BFBFBF}5763.26   & 0.00           &                         & \cellcolor[HTML]{BFBFBF}5763.26  & 5763.26                           & \cellcolor[HTML]{BFBFBF}5763.26   & \cellcolor[HTML]{BFBFBF}5763.26   \\
d1655-3                                         & 23921.90                                  & \cellcolor[HTML]{F2F2F2}25229.30 & 25403.26                          & \cellcolor[HTML]{D9D9D9}23921.90  & \cellcolor[HTML]{BFBFBF}23853.70  & \textbf{-0.29} &                         & \cellcolor[HTML]{F2F2F2}25635.98 & 25804.06                          & \cellcolor[HTML]{BFBFBF}24149.72  & \cellcolor[HTML]{D9D9D9}24179.13  \\
d1655-5                                         & 16512.20                                  & \cellcolor[HTML]{F2F2F2}17181.20 & 17502.88                          & \cellcolor[HTML]{D9D9D9}16512.20  & \cellcolor[HTML]{BFBFBF}16366.40  & \textbf{-0.88} &                         & \cellcolor[HTML]{F2F2F2}17454.32 & 17824.61                          & \cellcolor[HTML]{BFBFBF}16754.81  & \cellcolor[HTML]{D9D9D9}17025.55  \\
d1655-10                                        & 11320.10                                  & \cellcolor[HTML]{F2F2F2}11660.00 & 11814.34                          & \cellcolor[HTML]{BFBFBF}11320.10  & \cellcolor[HTML]{D9D9D9}11430.20  & 0.97           && \cellcolor[HTML]{F2F2F2}11816.04 & 11971.60                          & \cellcolor[HTML]{BFBFBF}11528.33  & \cellcolor[HTML]{D9D9D9}11761.92  \\
d1655-20                                        & 9598.94                                   & \cellcolor[HTML]{BFBFBF}9598.94  & 9910.12                           & \cellcolor[HTML]{D9D9D9}9627.28   & \cellcolor[HTML]{BFBFBF}9598.94   & 0.00           & & \cellcolor[HTML]{D9D9D9}9607.73  & 10172.24                          & \cellcolor[HTML]{F2F2F2}9669.97   & \cellcolor[HTML]{BFBFBF}9602.37   \\
u2152-3                                         & 22127.80                                  & 24207.40                         & \cellcolor[HTML]{F2F2F2}23295.73  & \cellcolor[HTML]{D9D9D9}22127.80  & \cellcolor[HTML]{BFBFBF}22010.50  & \textbf{-0.53} &                         & 24747.01                         & \cellcolor[HTML]{F2F2F2}23746.41  & \cellcolor[HTML]{D9D9D9}22629.25  & \cellcolor[HTML]{BFBFBF}22120.16  \\
u2152-5                                         & 14094.00                                  & 15055.10                         & \cellcolor[HTML]{F2F2F2}14778.56  & \cellcolor[HTML]{D9D9D9}14094.00  & \cellcolor[HTML]{BFBFBF}13776.60  & \textbf{-2.25} &                         & 15394.85                         & \cellcolor[HTML]{F2F2F2}15236.58  & \cellcolor[HTML]{D9D9D9}14442.21  & \cellcolor[HTML]{BFBFBF}13843.31  \\
u2152-10                                        & 8332.12                                   & \cellcolor[HTML]{F2F2F2}8624.61  & 8763.71                           & \cellcolor[HTML]{D9D9D9}8332.12   & \cellcolor[HTML]{BFBFBF}7981.80   & \textbf{-4.20} &                         & \cellcolor[HTML]{F2F2F2}8780.91  & 9018.30                           & \cellcolor[HTML]{D9D9D9}8499.64   & \cellcolor[HTML]{BFBFBF}8101.72   \\
u2152-20                                        & 6171.89                                   & \cellcolor[HTML]{BFBFBF}6171.89  & 6605.48                           & \cellcolor[HTML]{D9D9D9}6253.35   & \cellcolor[HTML]{BFBFBF}6171.89   & 0.00           && \cellcolor[HTML]{D9D9D9}6225.82  & 6676.91                           & \cellcolor[HTML]{F2F2F2}6339.15   & \cellcolor[HTML]{BFBFBF}6171.89   \\
pr2392-3                                        & 130015.00                                 & 141627.00                        & \cellcolor[HTML]{F2F2F2}135763.02 & \cellcolor[HTML]{D9D9D9}130015.00 & \cellcolor[HTML]{BFBFBF}129213.00 & \textbf{-0.62} &                         & 143703.00                        & \cellcolor[HTML]{F2F2F2}137589.12 & \cellcolor[HTML]{D9D9D9}132228.70 & \cellcolor[HTML]{BFBFBF}129943.70 \\
pr2392-5                                        & 82408.50                                  & 88083.20                         & \cellcolor[HTML]{F2F2F2}87465.60  & \cellcolor[HTML]{D9D9D9}82408.50  & \cellcolor[HTML]{BFBFBF}80098.90  & \textbf{-2.80} &                         & 89582.83                         & \cellcolor[HTML]{F2F2F2}88179.42  & \cellcolor[HTML]{D9D9D9}84448.64  & \cellcolor[HTML]{BFBFBF}80581.39  \\
pr2392-10                                       & 49033.60                                  & 51085.30                         & \cellcolor[HTML]{F2F2F2}50514.84  & \cellcolor[HTML]{D9D9D9}49033.60  & \cellcolor[HTML]{BFBFBF}46131.80  & \textbf{-5.92} &                         & 52100.80                         & \cellcolor[HTML]{F2F2F2}50929.73  & \cellcolor[HTML]{D9D9D9}49985.84  & \cellcolor[HTML]{BFBFBF}46454.63  \\
pr2392-20                                       & 35325.30                                  & \cellcolor[HTML]{D9D9D9}35325.30 & 35999.41                          & \cellcolor[HTML]{F2F2F2}35455.50  & \cellcolor[HTML]{BFBFBF}34746.30  & \textbf{-1.64} &                         & 35709.02                         & \cellcolor[HTML]{F2F2F2}36546.00  & \cellcolor[HTML]{D9D9D9}36107.77  & \cellcolor[HTML]{BFBFBF}35285.25  \\
pcb3038-3                                       & 46994.60                                  & 51049.90                         & \cellcolor[HTML]{F2F2F2}48351.41  & \cellcolor[HTML]{D9D9D9}46994.60  & \cellcolor[HTML]{BFBFBF}46511.70  & \textbf{-1.03} &                         & 51582.38                         & \cellcolor[HTML]{F2F2F2}49081.79  & \cellcolor[HTML]{D9D9D9}47686.85  & \cellcolor[HTML]{BFBFBF}46853.08  \\
pcb3038-5                                       & 29223.00                                  & 31140.20                         & \cellcolor[HTML]{F2F2F2}30089.85  & \cellcolor[HTML]{D9D9D9}29223.00  & \cellcolor[HTML]{BFBFBF}28521.30  & \textbf{-2.40} &                         & 31495.59                         & \cellcolor[HTML]{F2F2F2}30603.78  & \cellcolor[HTML]{D9D9D9}29864.61  & \cellcolor[HTML]{BFBFBF}28678.01  \\
pcb3038-10                                      & 16031.70                                  & 16949.90                         & \cellcolor[HTML]{F2F2F2}16878.69  & \cellcolor[HTML]{D9D9D9}16031.70  & \cellcolor[HTML]{BFBFBF}15462.10  & \textbf{-3.55} &                         & 17450.44                         & \cellcolor[HTML]{F2F2F2}16645.10  & \cellcolor[HTML]{D9D9D9}16509.95  & \cellcolor[HTML]{BFBFBF}15545.04  \\
pcb3038-20                                      & 10769.60                                  & 10835.00                         & \cellcolor[HTML]{F2F2F2}10827.78  & \cellcolor[HTML]{D9D9D9}10769.60  & \cellcolor[HTML]{BFBFBF}10375.70  & \textbf{-3.66} &                         & \cellcolor[HTML]{F2F2F2}11004.40 & 11196.47                          & \cellcolor[HTML]{D9D9D9}10961.26  & \cellcolor[HTML]{BFBFBF}10530.69  \\
fl3795-3                                        & 10927.40                                  & \cellcolor[HTML]{F2F2F2}11971.00 & 12290.18                          & \cellcolor[HTML]{BFBFBF}10927.40  & \cellcolor[HTML]{D9D9D9}11106.00  & 1.63           &                         & \cellcolor[HTML]{F2F2F2}12815.54 & 13022.12                          & \cellcolor[HTML]{BFBFBF}11321.19  & \cellcolor[HTML]{D9D9D9}11889.81  \\
fl3795-5                                        & 7715.36                                   & \cellcolor[HTML]{F2F2F2}7923.71  & 8151.43                           & \cellcolor[HTML]{D9D9D9}7715.36   & \cellcolor[HTML]{BFBFBF}7615.50   & \textbf{-1.29} &                         & \cellcolor[HTML]{D9D9D9}8610.84  & \cellcolor[HTML]{F2F2F2}8657.18   & \cellcolor[HTML]{BFBFBF}8014.97   & 8750.10                           \\
fl3795-10                                       & 5763.26*                                  & \cellcolor[HTML]{BFBFBF}5763.26  & 5824.14                           & \cellcolor[HTML]{D9D9D9}5764.85   & \cellcolor[HTML]{F2F2F2}5805.59   & 0.73           && \cellcolor[HTML]{D9D9D9}5823.89  & \cellcolor[HTML]{F2F2F2}5990.03   & \cellcolor[HTML]{BFBFBF}5810.07   & 6416.12                           \\
fl3795-20                                       & 5763.26*                                  & \cellcolor[HTML]{BFBFBF}5763.26  & \cellcolor[HTML]{BFBFBF}5763.26   & \cellcolor[HTML]{BFBFBF}5763.26   & \cellcolor[HTML]{BFBFBF}5763.26   & 0.00           &                         & \cellcolor[HTML]{BFBFBF}5763.26  & 5766.51                           & \cellcolor[HTML]{F2F2F2}5763.74   & \cellcolor[HTML]{BFBFBF}5763.26   \\
fnl4461-3                                       & 62016.70                                  & 66903.70                         & \cellcolor[HTML]{F2F2F2}64140.70  & \cellcolor[HTML]{D9D9D9}62016.70  & \cellcolor[HTML]{BFBFBF}61643.60  & \textbf{-0.60} &                         & 67971.34                         & \cellcolor[HTML]{F2F2F2}65268.41  & \cellcolor[HTML]{D9D9D9}62894.65  & \cellcolor[HTML]{BFBFBF}61955.88  \\
fnl4461-5                                       & 38265.50                                  & 40721.20                         & \cellcolor[HTML]{F2F2F2}39839.40  & \cellcolor[HTML]{D9D9D9}38265.50  & \cellcolor[HTML]{BFBFBF}37382.40  & \textbf{-2.31} &                         & 41777.11                         & \cellcolor[HTML]{F2F2F2}39839.40  & \cellcolor[HTML]{D9D9D9}38935.22  & \cellcolor[HTML]{BFBFBF}37698.19  \\
fnl4461-10                                      & 20671.60                                  & \cellcolor[HTML]{F2F2F2}22041.50 & 22145.65                          & \cellcolor[HTML]{D9D9D9}20671.60  & \cellcolor[HTML]{BFBFBF}19499.40  & \textbf{-5.67} &                         & \cellcolor[HTML]{F2F2F2}22891.45 & 22954.31                          & \cellcolor[HTML]{D9D9D9}20996.25  & \cellcolor[HTML]{BFBFBF}19660.79  \\
fnl4461-20                                      & 12347.80                                  & \cellcolor[HTML]{F2F2F2}12630.10 & 12789.54                          & \cellcolor[HTML]{D9D9D9}12347.80  & \cellcolor[HTML]{BFBFBF}11275.80  & \textbf{-8.68} &                         & 13046.38                         & \cellcolor[HTML]{F2F2F2}12987.65  & \cellcolor[HTML]{D9D9D9}12542.14  & \cellcolor[HTML]{BFBFBF}11444.20  \\
rl5915-3                                        & 193879.00                                 & 213864.00                        & \cellcolor[HTML]{F2F2F2}210056.49 & \cellcolor[HTML]{D9D9D9}193879.00 & \cellcolor[HTML]{BFBFBF}190121.00 & \textbf{-1.94} &                         & 226819.75                        & \cellcolor[HTML]{F2F2F2}215466.06 & \cellcolor[HTML]{D9D9D9}198796.00 & \cellcolor[HTML]{BFBFBF}191662.90 \\
rl5915-5                                        & 120418.00                                 & 133457.00                        & \cellcolor[HTML]{F2F2F2}125537.65 & \cellcolor[HTML]{D9D9D9}120418.00 & \cellcolor[HTML]{BFBFBF}116331.00 & \textbf{-3.39} &                         & 145173.07                        & \cellcolor[HTML]{F2F2F2}132524.74 & \cellcolor[HTML]{D9D9D9}124718.05 & \cellcolor[HTML]{BFBFBF}116977.65 \\
rl5915-10                                       & 66329.40                                  & 76585.20                         & \cellcolor[HTML]{F2F2F2}70853.30  & \cellcolor[HTML]{D9D9D9}66329.40  & \cellcolor[HTML]{BFBFBF}61375.00  & \textbf{-7.47} &                         & 84459.02                         & \cellcolor[HTML]{F2F2F2}71353.13  & \cellcolor[HTML]{D9D9D9}67961.45  & \cellcolor[HTML]{BFBFBF}62271.92  \\
rl5915-20                                       & 43121.00                                  & 48958.50                         & \cellcolor[HTML]{F2F2F2}44716.69  & \cellcolor[HTML]{D9D9D9}43121.00  & \cellcolor[HTML]{BFBFBF}39227.50  & \textbf{-9.03} &                         & 60306.22                         & \cellcolor[HTML]{F2F2F2}46080.96  & \cellcolor[HTML]{D9D9D9}44373.90  & \cellcolor[HTML]{BFBFBF}39739.66  \\
\hline
Average                                         & 36758.87                                  & 35077.55                         & 34076.09                          & 32450.03                          & 31608.39                          & -              &                         & 36713.40                         & 34801.53                          & 33158.00                          & 31951.21   \\
\hline               
\end{tabular}
\end{tiny}
\end{center}
%\vskip -0.10in
%\end{sidewaystable}
\end{table*}
\renewcommand{\baselinestretch}{1.0}\large\normalsize

\end{document}